\documentclass{article}

\usepackage[table, dvipsnames]{xcolor}
\usepackage{colortbl}

\usepackage{microtype}
\usepackage{graphicx}
\usepackage{subfigure}
\usepackage{booktabs} % for professional tables

\usepackage{hyperref}

\usepackage[accepted]{icml2025}

\usepackage{amsmath}
\usepackage{amssymb}
\usepackage{mathtools}
\usepackage{amsthm}

\usepackage[capitalize,noabbrev]{cleveref}

\theoremstyle{plain}
\newtheorem{theorem}{Theorem}[section]

\newtheorem{lemma}[theorem]{Lemma}

\theoremstyle{definition}
\newtheorem{definition}[theorem]{Definition}
\newtheorem{assumption}[theorem]{Assumption}
\theoremstyle{remark}

\usepackage[textsize=tiny]{todonotes}

\usepackage{nccmath}    % \medmath
\usepackage{amssymb}    % xmark & cmark
\usepackage{pifont}     % xmark & cmark
\usepackage{array,booktabs}
\usepackage{svg}

\usepackage{float}                          % For [H]
\usepackage{multirow}
\usepackage[flushleft]{threeparttable}    % For threeparttable
\usepackage{booktabs}                     % For \toprule, \midrule, \cmidrule, etc
\usepackage{makecell}
\usepackage{adjustbox}
\usepackage{caption}
\usepackage{subcaption} % for \subfigure (or use subfig)

\usepackage{amssymb}
\usepackage{tikz}

\def\tabref#1{Table~\ref{#1}}
\def\eqref#1{\cref{#1}}
\def\secref#1{Section~\ref{#1}}

\DeclareMathOperator*{\argmax}{arg\,max}
\DeclareMathOperator*{\argmin}{arg\,min}
\newcommand{\err}[2]{\epsilon_{#1}(#2)}

\newcommand{\bx}{\mathbf{x}}
\newcommand{\bz}{\mathbf{z}}

\newcommand{\Sim}{\boldsymbol{s}}
\newcommand{\bpi}{{\boldsymbol{\pi}}}

\newcommand{\KLD}[2]{\text{KL}(#1|#2)}

\newcommand{\pred}{\hat{p}}
\newcommand{\cocoamix}{\hat{p}_{\mathcal{T}}^\bpi}
\newcommand{\lossprompt}{\mathcal{L}_\text{prompt}}

\newcommand{\vtwo}[1]{#1}

\newcommand{\EDITED}[1]{#1}

\newcommand{\version}{1} % 1 == orig;  2 == new(2 phase)
\newcommand{\vcase}[2]{%
  \ifthenelse{\equal{\version}{1}}{%
  {#1}} % {\textcolor{magenta}{#1}}}%
  {\textcolor{blue}{#2}}%
}

\icmltitlerunning{CoCoA-Mix: Confusion-and-Confidence-Aware Mixture Model for Context Optimization}

\begin{document}

\twocolumn[
% \icmltitle{Submission and Formatting Instructions for \\
%            International Conference on Machine Learning (ICML 2025)}
% \icmltitle{From Mastery to Versatility: \\
%            Confusion and Confidence Aware Context Optimization}
\icmltitle{CoCoA-Mix: Confusion-and-Confidence-Aware\\
           Mixture Model for Context Optimization}



\icmlsetsymbol{equal}{*}

\begin{icmlauthorlist}
\icmlauthor{Dasol Hong}{url}
\icmlauthor{Wooju Lee}{url}
\icmlauthor{Hyun Myung}{url}
% \icmlauthor{Firstname2 Lastname2}{equal,yyy,comp}
% \icmlauthor{Firstname3 Lastname3}{comp}
\end{icmlauthorlist}

\icmlaffiliation{url}{Urban Robotics Lab, School of Electrical Engineering, Korea Advanced Institute of Science and Technology, Republic of Korea}
\icmlcorrespondingauthor{Dasol Hong}{ds.hong@kaist.ac.kr}
\icmlcorrespondingauthor{Wooju Lee}{dnwn24@kaist.ac.kr}
\icmlcorrespondingauthor{Hyun Myung}{hmyung@kaist.ac.kr}

\icmlkeywords{Machine Learning, ICML}

\vskip 0.3in
]



\printAffiliationsAndNotice{}  % leave blank if no need to mention equal contribution

\begin{abstract}
% This document provides a basic paper template and submission guidelines.
% Abstracts must be a single paragraph, ideally between 4--6 sentences long.
% Gross violations will trigger corrections at the camera-ready phase.

Prompt tuning, which adapts vision-language models by freezing model parameters and optimizing only the prompt, has proven effective for task-specific adaptations.
The core challenge in prompt tuning is improving specialization for a specific task and generalization for unseen domains. 
However, frozen encoders often produce misaligned features, leading to confusion between classes and limiting specialization.
To overcome this issue, we propose a confusion-aware loss (CoA-loss) that improves specialization by refining the decision boundaries between confusing classes.
Additionally, we mathematically demonstrate that a mixture model can enhance generalization without compromising specialization.
% This is achieved using confidence-aware temperature (CoA-temp), which adjusts the weights of each prediction in the mixture model based on its confidence within the class domains.
This is achieved using \vtwo{confidence-aware weights (CoA-weights)}, which adjust the weights of each prediction in the mixture model based on its confidence within the class domains.
Extensive experiments show that CoCoA-Mix, a mixture model with CoA-loss and \vtwo{CoA-weights}, outperforms state-of-the-art methods by enhancing specialization and generalization.
Our code is publicly available at \href{https://github.com/url-kaist/CoCoA-Mix}{https://github.com/url-kaist/CoCoA-Mix}.

\end{abstract}
\section{Introduction}
\label{sec:introduction}

\vtwo{Pre-trained} vision-language models (VLMs) have achieved remarkable results in diverse downstream tasks, such as image classification~\cite{fu2022cma}, object detection~\cite{zhong2022regionclip, gu2021open}, and visual question answering~\cite{cho2021unifying, lin2022revive}. These models align visual and textual embeddings through extensive pre-training on large-scale datasets, enabling remarkable zero-shot capabilities.
However, their reliance on generic embeddings often limits their effectiveness in task-specific applications.

\begin{figure}[!t]
    \centering  
    \subfigure[]
    {\label{fig:figure1_a}\includegraphics[width=80mm]{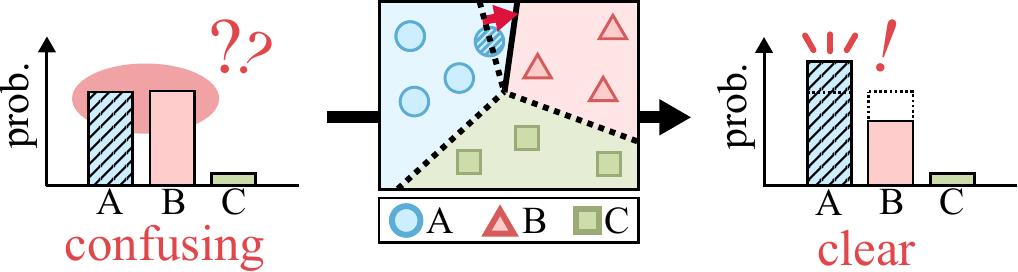}}
    \subfigure[]{\label{fig:figure1_b}\includegraphics[width=80mm]{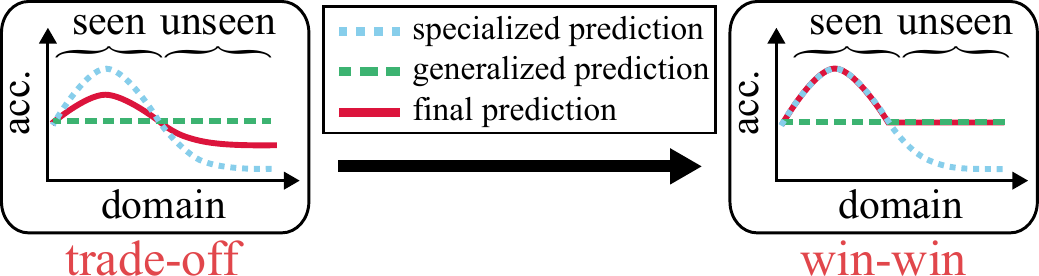}}
    \caption{
    (a) Effect of CoA-loss. 
    The left and right sides show probabilities for the hatched sample on the decision boundary. 
    % In the middle figure, dashed lines and background colors indicate the current and ideal decision boundaries, respectively.
    In the middle figure, dashed and solid lines indicate the current and updated decision boundaries.
    CoA-loss assigns higher weights to this confusing sample, enhancing specialization in prompt tuning.
    (b) Effect of \vtwo{CoA-weights}. 
    The left and right sides show performance across domains. \vtwo{CoA-weights increase} confidence in the specialized prediction for seen domains while reducing it for unseen domains, preserving both specialization and generalization.
    }\label{fig:figure1}
\end{figure}

To mitigate this limitation, prompt engineering has emerged as a practical way to enhance performance on a specific task~\cite{wang2021simvlm, wang2022ofa}. % feng2023promptmagician, oppenlaender2024prompting}.
The method involves manually designing a task-specific text template for the model input.
For example, in image classification, hand-crafted prompts such as \texttt{"a photo of a [CLASS], a type of flower"} help associate visual embeddings with the correct class labels, such as ``hibiscus" or ``sword lily."~\cite{zhou2022learning}
Although effective, the manual process is labor-intensive and requires domain expertise, limiting scalability across diverse tasks. 

Prompt tuning has become a scalable alternative to manual prompt engineering by replacing hand-crafted prompts with learnable prompts~\cite{zhou2022learning, zhu2023prompt, pmlr-v235-zhou24s, zhang2024amend}. % ~\cite{zhou2022learning, zhu2023prompt, khattak2023maple, pmlr-v235-zhou24s, zhang2024amend}. 
This method allows VLMs to effectively align textual embeddings with the corresponding visual embeddings by optimizing the prompts for a specific task while freezing the model parameters.

However, the existing methods have two limitations.
% First, the frozen visual encoder often fails to capture task-specific features.
% As a result, it often leads to confusing predictions and fails to distinguish between different classes.
First, existing methods do not explicitly address confusing cases arising from the frozen visual encoder. The frozen visual encoder often fails to capture task-specific features, struggling to distinguish between different classes.
However, most existing methods rely on standard cross-entropy loss, which is ineffective in handling confusing cases and consequently limits their specialization~\cite{zhou2022learning, zhou2022conditional}. Second, achieving generalization to unseen domains is crucial in prompt tuning, but current methods often sacrifice specialization to improve generalization.  
Most studies %~\cite{zhu2023prompt, yao2023visual, khattak2023maple, yao2024tcp} 
inherently assume specialization and generalization are competing objectives.
As a result, the trade-off problem remains an open challenge.

To overcome these limitations, we propose a confusion-and-confidence-aware mixture model (CoCoA-Mix), which combines confusion-aware loss (CoA-loss) and \vtwo{confidence-aware weights (CoA-weights)}.
We first introduce a mixture model that combines \vtwo{predictions} from individual prompts, providing a theoretical framework to analyze prompt tuning in terms of specialization and generalization.
Building on this theoretical insight, CoA-loss improves specialization by applying larger gradient to confusing cases, refining decision boundaries as shown in~\cref{fig:figure1_a}.
Then, \vtwo{CoA-weights achieve} generalization without compromising specialization by scaling mixture model weights based on the confidence of individual prompts across class domains, as shown in~\cref{fig:figure1_b}.
As a result, our method enhances both specialization and generalization by ensuring that the error of the mixture model remains lower than the minimum error of individual prompts.
Main contributions are as follows:

\begin{itemize}
    \item We provide a mathematical framework demonstrating that specialization and generalization can be improved simultaneously.
    \item We propose a CoCoA-Mix framework, consisting of CoA-loss and \vtwo{CoA-weights}. CoA-loss boosts specialization by improving classification for confusing cases, while \vtwo{CoA-weights improve} generalization by adjusting the weights of individual prompts in the mixture model based on their confidence over class domains.
    \item The proposed method achieves average harmonic mean improvements of \vcase{$15.28\%$ and $3.28\%$}{$?\%$ and $?\%$} over zero-shot CLIP in base-to-new generalization and cross-dataset transfer, respectively;
    %%% v1
    % it also improves the average accuracy in few-shot class-incremental learning by $4.4\%p$.
    %%% v2
    it also improves the average accuracy in few-shot class-incremental learning by $5.6\%p$.
    % \vcase{$4.4\%p$}{$?\%p$}.
\end{itemize}

\section{Related Work}
\label{sec:related_works}

\newcommand{\red}[1]{\textcolor{red}{#1}}

%%%%%%%%%%%%%%%%%%%%%%%%%%%%%%%%%%%%%%%%%%%%%%%%%%%
\subsection{Textual Prompt Tuning}

Prompt tuning has emerged as a method to reduce the reliance on human expertise while improving the performance of VLMs on specific tasks.
The method can be categorized into visual prompt tuning~\cite{jia2022visual, bahng2022exploring}, textual prompt tuning~\cite{zhou2022learning, zhou2022conditional, zhu2023prompt, yao2023visual, zhang2024dept}, and visual-textual prompt tuning~\cite{khattak2023maple, khattak2023self}. 
% In this paper, we focus on textual prompt tuning.
% Textual prompt tuning replaces hand-crafted prompts with learnable prompts and optimizes them in the training domain while keeping the model parameters frozen.
\vtwo{
In this paper, we focus on textual prompt tuning, which replaces hand-crafted prompts with learnable prompts and optimizes them using few-shot training data, while keeping the model parameters frozen.
}

\subsection{Loss for Prompt Tuning}
% \subsection{Textual Prompt Tuning}
%
%\paragraph{Specialization in Prompt Tuning}
%
\vtwo{In textual prompt tuning,} % In prompt tuning, 
the frozen encoders map inputs to generic embeddings, which can lead to misaligned vision-text embeddings for specific tasks. 
A common strategy to mitigate this issue is to employ a standard cross-entropy loss that aligns textual prompts with their corresponding visual embeddings.
CoOp~\cite{zhou2022learning} pioneered textual prompt tuning by optimizing prompts for specific tasks using cross-entropy loss.
% However, cross-entropy loss can be less effective at handling confusing samples, where different classes are predicted with similar probabilities, limiting its ability to distinguish them.
Subsequent works employed regularization via hand-crafted prompts, which enhanced specialization by preventing prompts from learning unintended patterns~\cite{zhu2023prompt, yao2023visual, zhang2024dept}.
However, such regularization methods may prevent prompts from fully capturing task-specific patterns, limiting specialization in complex tasks.
% In this paper, we aim to enhance specialization through CoA-loss, enabling the model to clearly distinguish confusing classes.
% \paragraph{Specialization-Generalization Trade-offs in Prompt Tuning}
%
% Achieving generalization in unseen domains without compromising specialization is another critical problem in prompt tuning.
\vtwo{Other approaches~\cite{khattak2023maple, zhang2024dept} introduce additional network parameters to enhance the representational capacity of the model.}
% Previous approaches~\cite{khattak2023maple, zhang2024dept} address the trade-off by introducing additional network parameters to enhance the representational capacity of the model.
MaPLe~\cite{khattak2023maple} leverages visual-language interactions via a coupling function, while 
DePT~\cite{zhang2024dept} employs a dual-head architecture to decouple task-specific and task-shared knowledge into separate feature spaces.
Although these methods improve generalization, they increase the number of learnable parameters, leading to overfitting when training data is scarce and limiting scalability.
In contrast, we aim to enhance both specialization and generalization without introducing additional network components.  
% We introduce a mixture model to mathematically analyze prompt tuning performance from specialization and generalization perspectives, demonstrating that both can be improved simultaneously.  
% This is further supported by CoA-temp, which adjusts the confidence of prompt predictions based on class domains.

\subsection{Mixture of Prompts}

\vtwo{
Prompt ensembling has been studied to improve generalization in VLMs. 
Allingham et al.~\yrcite{allingham2023simple} propose zero-shot prompt ensembling (ZPE), which automatically assigns weights to hand-crafted prompts from a large pool, leading to improved zero-shot accuracy.
Lu et al.~\yrcite{lu2024beyond} extend this to model-level ensembles suited to varying resource settings, % though at the cost of multiple forward passes during inference.
but this requires multiple forward passes at inference time.
Despite their generalization benefits, neither approach explicitly addresses the challenge of achieving task-specific specialization.
In contrast, our method jointly improves specialization and generalization, while remaining computationally efficient by avoiding multiple forward passes.
}

\begin{figure*}[!t]
    \centering
    \vcase{
    \includegraphics[width=0.85\textwidth]{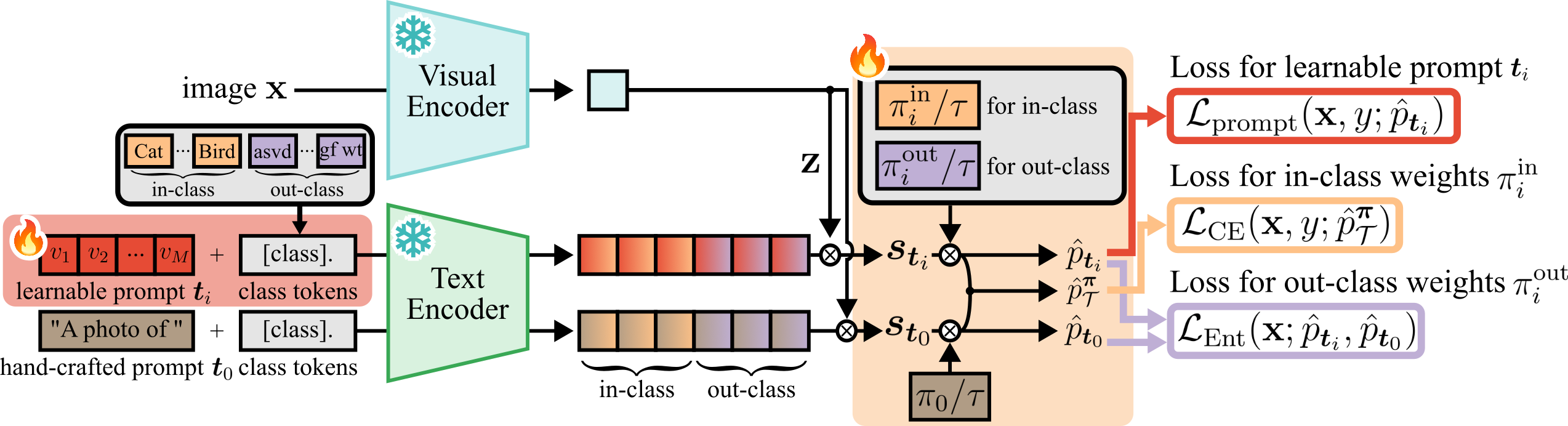}
    }{
    \includegraphics[width=\textwidth]{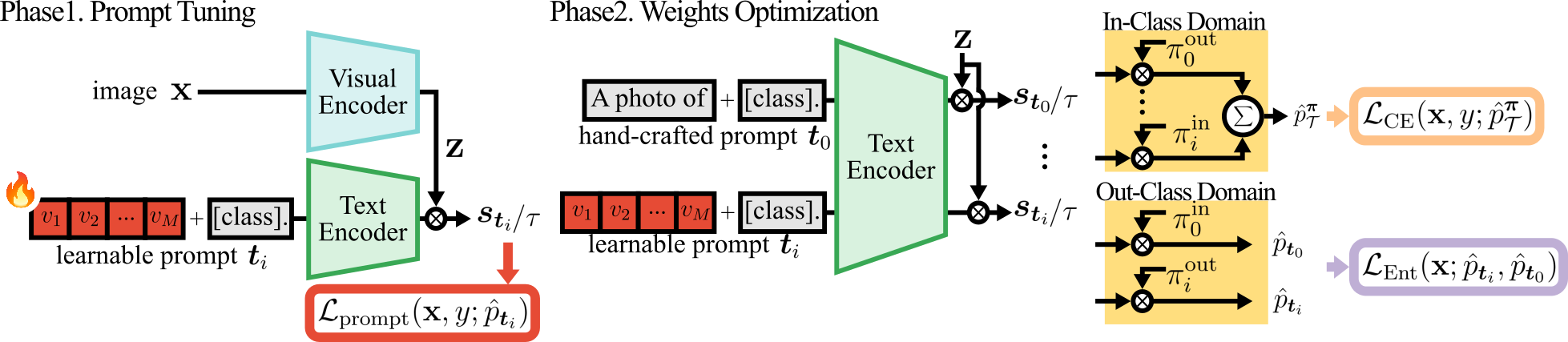}
    }
    \caption{
    CoCoA-Mix framework integrates confusion-aware loss (CoA-loss) for specialization and \vtwo{confidence-aware weights (CoA-weights)} % $\pi_i$ 
    for generalization, ensuring performance improvements without trade-offs. 
    The learnable prompt $\boldsymbol{t}_i$ is optimized with CoA-loss to specialize in distinguishing confusing classes within the training domain.
    % \vtwo{CoA-weights} adjusts prediction confidence by increasing $\tau_i^\text{in}$ for in-class and decreasing $\tau_i^\text{out}$ for out-class.
    CoA-weights adjusts prediction confidence by increasing $\pi_i^\text{in}$ for in-class and decreasing $\pi_i^\text{out}$ for out-class.
    At inference, the specialized predictions $\hat{p}_{\mathbf{t}_i}$, adjusted via \vtwo{CoA-weights}, are combined with the generalized predictions $\hat{p}_{\mathbf{t}_0}$ to ensure generalization while preserving specialization.
    }
    \label{fig:overview}
\end{figure*}

\section{Proposed Method}
\label{sec:method}
%%%%%%%%%%%%%%%%%%%%%%%%%%%%%%%%%
%%% Version 1.4. (2025.01.01) %%%
%%%%%%%%%%%%%%%%%%%%%%%%%%%%%%%%%
\newcommand{\Exy}[1]{\mathbb{E}_{(\bx, y)\sim\mathcal{D}_{#1}}}
\newcommand{\bt}{{\boldsymbol{t}}}

%%%%%%%%%%%%%%%%%%%%%%%%%%%%%%%%%%%%%%%%%%%%%%%%%%%
\subsection{Preliminary}

\newcommand{\softmaxt}[3]{\frac{\exp\left(#1(#3)/{#2}\right)}{\sum_{l'\in\mathcal{Y}}\exp\left(#1(l')/{#2}\right)}}
\newcommand{\fvis}{f_{\text{vis}}}
\newcommand{\ftxt}{f_{\text{txt}}}

In downstream tasks, CLIP employs visual and textual encoders to map images and textual prompts into a shared embedding space.
The textual prompts $\bt$ can be generated as hand-crafted prompts by embedding class labels into predefined templates, such as \texttt{"a photo of a [CLASS]."}
With $\bt$, the probability of the image belonging to the class $l$, $\pred_\bt(l)$, is defined as follows:
\begin{equation}
    \pred_\bt(l) = \softmaxt{\Sim_\bt}{\tau}{l},
\end{equation}
\noindent where $\tau$ is the temperature scale, $\mathcal{Y}$ represents the set of classes, and $\Sim_\bt(l)$ denotes the cosine similarity between the visual and textual embeddings of the class $l$ generated by $\bt$.

% \subsection{Expected Error of the Mixture Model}
\vtwo{
% \subsection{Specialization and Generalization Analysis with Mixture Models}
\subsection{Decomposing Specialization and Generalization}
}

The expected error $\err{T}{\pred}$ of a predictive distribution $\pred$ in an arbitrary target domain $\mathcal{D}_T$ is defined using the Kullback-Leibler (KL) divergence as follows:
%% Equation 1 : Expected Error %%
\begin{equation}\label{eq:expected_error} 
    \err{T}{\pred} 
    % = \Exy{T}\left[ -\log{\pred(y|\bx)} \right],
    = \Exy{T}\left[ -\log{\pred(y)} \right],
\end{equation}
where $y$ is the ground-truth label for the image $\bx$.

We introduce a mixture model to derive an upper bound for the error in the target domain, providing a framework for analyzing specialization and generalization.

%% Definition 3.1 : Mixture Model %%
\begin{definition}[Mixture model] 

Let $K+1$ different prompts be given by $\mathcal{T}=\{\bt_0,\bt_1,\cdots,\bt_K\}$, and let $\bpi=\{\pi_0,\pi_1,\cdots,\pi_K\}$ denote a set of non-negative weights satisfying $\sum_{i=0}^K{\pi_i}=1$.
The mixture model $\cocoamix$ is defined as a weighted combination of the individual prompts:
\begin{equation}
    \cocoamix (l)
    = \softmaxt{\sum_{i=0}^K\pi_i\Sim_{\bt_i}}{\tau}{l}.
\end{equation}
\end{definition}

%% Theorem 3.2 : Expected error of the mixture model %%
\begin{theorem}
\label{thm:expected_error_of_mixture_hypothesis}
The expected error of the mixture model $\cocoamix$ can be bounded as follows:
\begin{equation}
    \err{T}{\cocoamix} 
    \leq \sum_{i=0}^K \pi_i \err{T}{\pred_{\bt_i}}.
\end{equation}
The proof is provided in~\secref{sec:appendix-proof-main_them}.
\end{theorem}

\vtwo{
\begin{lemma}
\label{lemma:lemma33}
Let the class set of the target domain $\mathcal{D}_T$ be partitioned into $K+1$ disjoint subsets, with corresponding sub-domains $\mathcal{D}_{T_0},\mathcal{D}_{T_1},\cdots,\mathcal{D}_{T_K}$, such that $\mathcal{D}_T=\bigsqcup_{i=0}^K\mathcal{D}_{T_i}$. Then, the expected error of $\cocoamix$ is given by:
\begin{equation}
    \epsilon_T(\cocoamix) = \sum_{i=0}^K{
\lambda_i \epsilon_{T_i}(\cocoamix)
},
\end{equation}
\noindent where $\lambda_i=\Pr_{(\bx,y)\sim\mathcal{D}_T}\left[(\bx,y)\in\mathcal{D}_{T_i}\right]$ denotes the probability that a sample from the target domain $\mathcal{D}_T$ belongs to the sub-domain $\mathcal{D}_{T_i}$, satisfying $\sum_{i=0}^K\lambda_i=1$.
Based on \cref{thm:expected_error_of_mixture_hypothesis}, the error of the mixture model in the arbitrary target domain can be upper-bounded as follows:
\begin{equation}\label{eq:lemma33}
    \epsilon_T(\cocoamix) \leq \sum_{i=0}^K{
\lambda_i \left(
\pi_{i}^\text{in}
\underbrace{
  \epsilon_{T_i}(\hat{p}_{\bt_i})
}_{\substack{\text{specialization}\\\text{error}}}
+
% \sum_{\substack{j=0 \\ j \ne i}}^K\pi_j^\text{out}
% \underbrace{
%   \epsilon_{T_i}(\hat{p}_{\bt_j})
% }_{\substack{\text{generalization} \\ \text{error}}}
\underbrace{
    \sum_{\substack{j=0 \\ j \ne i}}^K\pi_j^\text{out}
  \epsilon_{T_i}(\hat{p}_{\bt_j})
}_{\substack{\text{generalization} \\ \text{error}}}
\right)
},
\end{equation}
where $\pi_i^\text{in}$ denotes the mixing weight of the prompt $\bt_i$ for its own domain $\mathcal{D}_{T_i}$, and $\pi_j^\text{out}$ denotes the mixing weight of the prompt $\bt_j(j\ne i)$ when applied to the domain $\mathcal{D}_i$. Here, $\pi_i^\text{in} + \sum_{\substack{j=0 \\ j \ne i}}^K\pi_j^\text{out} = 1$.
\end{lemma}
}
\vtwo{
Let $\mathcal{Y}$ denote the set of all classes in the arbitrary target domain $\mathcal{D}_T$, which is partitioned as $\mathcal{Y}=\bigsqcup_{i=0}^K\mathcal{Y}_i$. 
% In most practical scenarios, training data is unavailable for certain subsets of classes. 
Specifically, we assume that labeled data is provided for $\mathcal{Y}_1,\cdots,\mathcal{Y}_K$, whereas no supervision is available for $\mathcal{Y}_0$. Due to the absence of labeled data in $\mathcal{D}_{T_0}$, the associated prompt $\bt_0$ cannot be specialized through training. Instead, we utilize a generalized hand-crafted prompt such as} $\verb|"a photo of a [CLASS]."|$

% The following subsections analyze the mixture model error $\err{T}{\pred_\bt^\bpi}$ from specialization and generalization perspectives.
% To simplify the discussion, we restrict our analysis to the case where $\bpi = \{1-\alpha, \alpha\}$, with $\alpha$ being a non-negative scalar.
% Let $\pred_\bh$ and $\pred_\bv$ denote generalized and specialized predictive distribution provided by a hand-crafted prompt $\bh$ and a learnable prompt $\bv$, respectively. 
% $\bh$ is designed to generalize across diverse target domains, while $\bv$ is specialized in the \EDITED{source} domain.
% For the target domain, the expected error $\err{T}{\pred_\bt^\bpi}$ of the mixture model is bounded as follows:
% \begin{equation}\label{eq:simple_error_of_mixture}
%     \err{T}{\pred_\bt^\bpi}
%     \leq (1-\alpha) \err{T}{\pred_\bh} + \alpha \err{T}{\pred_\bv}.
% \end{equation}

Building on this, we detail the proposed method for specialization and generalization in prompt tuning.
% Because the pre-defined $\bh$ determines $\err{T}{\pred_\bh}$, it cannot be reduced. Thus, \cref{sec:specialization} focuses on minimizing $\err{T}{\pred_\bv}$.
\vtwo{
In \cref{sec:specialization}, we propose confusion-aware loss (CoA-loss), which focuses on confusing cases in prompt tuning and effectively reduces the \textit{specialization} error $\err{T_i}{\hat{p}_{\bt_i}}$ for each sub-domain $\mathcal{D}_{T_i}$.
Subsequently, \cref{sec:generalization} enhances \textit{generalization} by optimizing confidence-aware weights (CoA-weights) based on the confidence of each prompt on the given class domain, tightening the upper bound of $\err{T}{\cocoamix}$.
As a result, the proposed method enhances specialization and generalization simultaneously.
\cref{fig:overview} shows the overall framework of the confusion-and-confidence-aware mixture model (CoCoA-Mix), which combines CoA-loss and CoA-weights.
}

%%%%%%%%%%%%%%%%%%%%%%%%%%%%%%%%%%%%%%%%%%%%%%%%%%%
\subsection{Confusion-Aware Loss for Specialization}\label{sec:specialization}

Let $\pred$ be a predictive distribution and $\mathcal{D}_S$ be the source domain. According to Nguyen et al.~\yrcite{nguyen2022kl}, the expected error $\err{T}{\pred}$ on the target domain $\mathcal{D}_T$ is bounded as follows:
\begin{equation}\label{eq:target_source_error_bound}
    \err{T}{\pred} 
    \leq \err{S}{\pred} +\frac{C}{\sqrt{2}}\sqrt{
        \KLD{p_T(\bz)}{p_S(\bz)} + \delta
    },
\end{equation}
\noindent where $\bz$ is defined as a visual embedding; $C$ is a constant that bounds $\log{\pred(l)}$, ensuring each class probability is at least $\exp(-C)$; 
$p_T$ and $p_S$ are the marginal distribution of $\bz$ for the target and source domains, respectively;
and $\delta$ denotes the conditional misalignment $\mathbb{E}_{p_T(\bx)}\left[ \KLD{p_T(y|\bx)}{p_S(y|\bx)} \right]$, which is typically small.

We focus on textual prompt tuning, which cannot optimize the visual embedding $\bz$ due to the frozen encoders.
% Thus, \cref{eq:target_source_error_bound} suggests that minimizing the source domain error $\err{S}{\pred}$ can reduce the target domain error $\err{T}{\pred}$, enabling error minimization in the mixture model via source domain \textit{specialization}.
Thus, \cref{eq:target_source_error_bound} suggests that minimizing $\err{S}{\pred}$ can reduce $\err{T}{\pred}$, enabling error minimization in the mixture model via source domain \textit{specialization}.
% 만일 domain도 추가한다면 KL term도 추가로 언급

% \vtwo{For the training dataset $\mathcal{D}_{S_i}$ corresponding to the target domain $\mathcal{D}_{T_i}$,} 
Most existing methods use standard cross-entropy, defined as follows, for the specialization in prompt tuning:
\EDITED{
\begin{equation}
    \begin{aligned}
        \mathcal{L}_\text{CE}(\bx,y; \pred_\bt)
        = -\log{\pred_\bt(y)}.
    \end{aligned}
\end{equation}
}
Cross-entropy assigns gradients based on class probabilities, assigning larger gradients to misclassified samples.
% While effective for handling misclassified samples, it relies only on individual probabilities without considering inter-class relationships.
While effective for misclassified cases, it relies only on individual probabilities and does not consider inter-class relationships, limiting its ability to handle class confusion.
Addressing confusing classes is crucial in prompt tuning with limited training data because they significantly impact the decision boundary.
% In prompt tuning with limited training data, addressing confusing classes where the model struggles to distinguish is crucial because these inter-class relationships significantly affect decision boundaries.
To overcome this limitation, we propose confusion-aware loss (CoA-loss), defined as follows:
\begin{equation}
    \begin{aligned}
        \mathcal{L}_\text{CoA}(\bx,y; \pred_\bt)
        =1 - \pred_\bt(y).
    \end{aligned}
\end{equation}
%
% which applies larger gradient updates to confusing cases. This loss penalizes predictions where correct and incorrect classes have similar probabilities, guiding the model to refine decision boundaries between confusing classes more precisely.
%
The overall loss $\lossprompt$ for optimizing the prompt $\bt$ is given by:
\begin{equation}
    \begin{aligned}
        \lossprompt(\bx,y; \pred_\bt)
        = \mathcal{L}_\text{CE} + w \mathcal{L}_\text{CoA},
    \end{aligned}
\end{equation}
\noindent where $w$ is a hyperparameter that balances the contribution of CoA-loss.

\begin{figure}[!t]
    \centering  
    \subfigure[]{\label{fig:grad_a}\includegraphics[width=40mm]{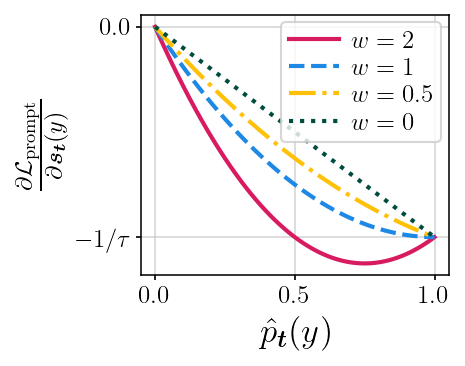}}
    \subfigure[]{\label{fig:grad_b}\includegraphics[width=40mm]{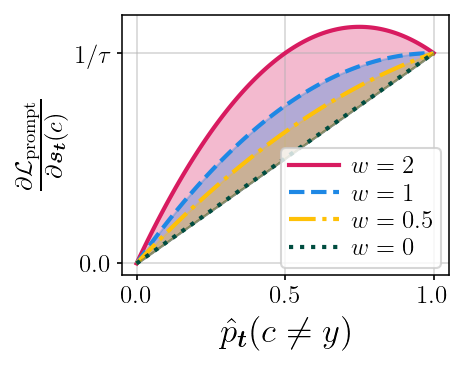}}
    \caption{Gradient component of $\lossprompt$ with respect to (a) $\Sim_\bt(c\neq y)$ and (b) $\Sim_\bt(y)$, where $w=0$ represents standard cross-entropy.
    }\label{fig:grad}
\end{figure}

To illustrate how CoA-loss refines the decision boundary between confusing classes, we analyze the gradients for the correct class $y$ and an incorrect class $c$.
The gradients of $\lossprompt$ with respect to the similarities $\Sim_\bt(y)$ and $\Sim_\bt(c\neq y)$ are as follows:

\begin{equation}\label{eq:grad}
    \begin{aligned}
        \frac{\partial{\lossprompt}}{\partial \Sim_\bt(y)} 
        = -\frac{1}{\tau}\left(1-\pred_\bt(y)\right)
        \left(1-w\pred_\bt(y)\right),\\
        \frac{\partial{\lossprompt}}{\partial \Sim_\bt(c\neq y)} 
        = \frac{1}{\tau}\pred_\bt(c)\left(1+w\pred_\bt(y)\right).\\
    \end{aligned}
\end{equation}

\cref{fig:grad} shows the gradients for classes $y$ and $c$. When $w=0$, the loss corresponds to standard cross-entropy. 
As implied by~\cref{eq:grad}, standard cross-entropy assigns gradients based on individual probabilities, ignoring inter-class relationships. In contrast, CoA-loss increases the gradient for $y$ when $\pred_\bt(y)$ approaches $0.5$ and for class $c$ when $\pred_\bt(c)$ nears $\pred_\bt(y)$, especially as $w$ increases.
These scenarios represent confusing cases where the correct class probability is uncertain or where the incorrect class probability is similar to the correct class, making them difficult to distinguish. CoA-loss thus induces larger gradient updates for these confusing cases.

%%%%%%%%%%%%%%%%%%%%%%%%%%%%%%%%%%%%%%%%%%%%%%%%%%%
\subsection{Confidence-Aware Weights for Generalization without Trade-Offs}\label{sec:generalization}

% \cref{eq:simple_error_of_mixture} suggests that optimizing $\alpha$ can reduce the expected error $\err{T}{\pred_\bt^\bpi}$ of the mixture model below the minimum error of individual predictions  $\pred_\bh$ and $\pred_\bv$ in the target domain.
\vtwo{
\cref{eq:lemma33} suggests that optimizing $\bpi$ can minimize the upper bound of the expected error $\err{T}{\cocoamix}$ of the mixture model.
}
Achieving this requires assigning a higher weight to the prediction with lower error in the target domain.
% For instance, if $\err{T}{\pred_\bv}$ is lower than $\err{T}{\pred_\bh}$, the weight $\alpha$ should be increased. 
% \vtwo{
% Specifically, for each accessible domain $\mathcal{D}_{T_i}$ ($i > 0$), we have a specialized prompt $\bt_i$ optimized via the CoA-Loss. 
% In this case, if $\err{T}{\pred_{\bt_i}}$ is lower than $\err{T}{\pred_{\bt_j}}$, the weight $\pi_i^\text{in}$ should be increased. 
% }
In this section, we propose confidence-aware weights (CoA-weights), which adjust the weight of each prediction based on the class domain.

\begin{assumption}\label{ass:divergence_class}
    \vtwo{
    The specialized prediction $\pred_{\bt_i}$ for $\mathcal{D}_{T_i}$ satisfies the following relationships:
    }
    \begin{equation}
        \begin{aligned}
            \err{T_i}{\pred_{\bt_i}} \leq \err{T_i}{\pred_{\bt_{j\neq i}}} 
            \quad\text{and}\quad
            \err{T_{j\neq i}}{\pred_{\bt_0}} \leq \err{T_{j\neq i}}{\pred_{\bt_i}}.
        \end{aligned}
    \end{equation}
    \vtwo{
    The first inequality reflects that a prediction $\pred_{\bt_i}$ optimized for a specific domain $\mathcal{D}_{T_i}$ always performs better than predictions $\pred_{\bt_{j\neq i}}$ made by prompts optimized for other domains.
    Conversely, the second inequality assumes that the generalized prediction $\pred_{\bt_0}$ is more effective for unseen classes.
    % This assumption aligns with the principle that a single prediction specialized for a task cannot excel at all tasks.
    This assumption aligns with the principle that a single prediction specialized for a task cannot excel at all tasks, a claim further supported by statistical evidence presented in~\cref{sec:appendix-ass_experiments}.
    }
\end{assumption}

\cref{eq:lemma33} and \cref{ass:divergence_class} suggest that minimizing the upper bound of $\err{T}{\cocoamix}$ requires increasing \vtwo{$\pi^\text{in}$} for in-classes and decreasing \vtwo{$\pi^\text{out}$} for out-classes.
\vtwo{
To achieve this, we propose CoA-weights which adjusts the weights $\pi_i^\text{in}$ and $\pi_i^\text{out}$ of $\pred_{\bt_i}$ in the mixture model.
}

%%% --- %%%
% \paragraph{Optimizing $\tau_\bv^\text{in}$ for In-Class Domains}
\paragraph{Optimizing $\pi_i^\text{in}$ for In-Class Domains}
% \subsubsection{Optimizing $\tau_\bv^\text{in}$ for In-Class Domains}

The \vtwo{weight $\pi_i^\text{in}$} for in-class domains is optimized by minimizing the cross-entropy loss of the mixture model over the training domain $\mathcal{D}_{S_i}$:
% \begin{equation}
%     \tau_\bv^\text{in} = \argmin_{\tau_\bv}{\mathbb{E}_{(\bx,y)\sim\mathcal{D}_{S_\bv}}\left[ \mathcal{L}_\text{CE}(\bx,y;\cocoamix) \right]}.
% \end{equation}
\vtwo{
\begin{equation}
    \pi_i^\text{in} = \argmin_{\pi_i^\text{in}}{\mathbb{E}_{(\bx,y)\sim\mathcal{D}_{S_i}}\left[ \mathcal{L}_\text{CE}(\bx,y;\cocoamix) \right]}.
\end{equation}
}

With this optimization, \vtwo{weight $\pi_i^\text{in}$} is increased when the specialized prediction \vtwo{$\pred_{\bt_i}$} outperforms the generalized prediction \vtwo{$\pred_{\bt_{j\neq i}}$} for the in-class domain and decreased otherwise. 
% Unlike regularization-based approaches~\cite{zhu2023prompt, yao2023visual}, optimizing the cross-entropy of the mixture model adjusts the temperature based on the performance of hand-crafted prompts in the target domain. As a result, it is less sensitive to incorrect prior knowledge in hand-crafted prompts. 
Consequently, the weight \vtwo{$\pi_i^\text{in}$} for in-classes is optimized. Further details on the cross-entropy effect in the mixture model are provided in~\cref{sec:appendix-effect_ce}.

%%% --- %%%
\paragraph{Optimizing $\pi_i^\text{out}$ for Out-Class Domains}
% \subsubsection{Optimizing $\tau_\bv^\text{out}$ for Out-Class Domains}

Because only in-classes are available during training, out-class set $\mathcal{Y}_i^\text{out}$ must be generated to optimize \vtwo{$\pi_i^\text{out}$}.
The out-class set can be generated by combining random strings or retrieving random words. 
According to~\cref{ass:divergence_class}, the temperature $\pi_i^\text{out}$ for out-class domains is optimized using entropy loss $\mathcal{L}_\text{Ent}$, which compares the entropy of $\pred_{\bt_i}$ and $\pred_{\bt_0}$ as follows:
% as follows:

% \begin{equation}
%     \tau_\bv^\text{out} = \argmin_{\tau_\bv}{\mathbb{E}_{(\bx,y)\sim\mathcal{D}_{S_\bv}}}\left[ 
%         \mathcal{L}_\text{Ent}(\bx;\pred_\bv,\pred_\bh)
%     \right],
% \end{equation}
\vtwo{
\begin{equation}
    \pi_i^\text{out} = \argmin_{\pi_i^\text{out}}{\mathbb{E}_{(\bx,y)\sim\mathcal{D}_{S_i}}}\left[ 
        \mathcal{L}_\text{Ent}(\bx;\pred_{\bt_i},\pred_{\bt_0})
    \right],
\end{equation}
}
% \begin{equation}
%     \mathcal{L}_\text{Ent}
%     = \max\left( 0, 
%         H\left(\pred_\bh\right)
%         -H\left(\pred_\bv\right)
%         + d 
%     \right),
% \end{equation}
\vtwo{
\begin{equation}
    \mathcal{L}_\text{Ent}
    = \max\left( 0, 
        H\left(\pred_{\bt_0}\right)
        -H\left(\pred_{\bt_i}\right)
        + d 
    \right),
\end{equation}
}

% \noindent where $d$ is a margin, and $H\left(\pred\right)$ is the normalized entropy of $\pred$ over the out-class set, defined as follows:
% \begin{equation} 
%     H\left(\pred\right) 
%     = \sum_{c\sim\mathcal{Y}_{S_\bv^\text{out}}}
%     \frac{
%         -\pred(c)\log \pred(c)
%     }{
%         \log {|\mathcal{Y}_{S_\bv^\text{out}}|}
%     }.
% \end{equation}

\noindent where $d$ is a margin and $H\left(\pred\right)$ is the normalized entropy of $\pred$ over the out-class set, 
% i.e., $H\left(\pred\right) 
%     = \sum_{c\sim\mathcal{Y}_{S_\bv^\text{out}}}
%     -\pred(c)\log \pred(c) / \log {|\mathcal{Y}_{S_\bv^\text{out}}|}$.
i.e., $H\left(\pred\right) 
    = \sum_{c\sim\mathcal{Y}_i^\text{out}}
    -\pred(c)\log \pred(c) / \log {|\mathcal{Y}_i^\text{out}|}$.

% $\tau_\bv^\text{out}$ is optimized using $\mathcal{L}_\text{Ent}$, which compares the entropy of $\pred_\bv$ and $\pred_\bh$.
The entropy loss $\mathcal{L}_\text{Ent}$ ensures that the entropy of $\pred_{\bt_{i>0}}$ exceeds that of $\pred_{\bt_0}$ by a margin of $d$.
Entropy measures uncertainty in the prediction, with higher values indicating lower confidence.
As a result, $\pi_i^\text{out}$ is optimized to make specialized predictions less confident than generalized ones. %, improving generalization. % without trade-offs.

% Entropy quantifies uncertainty in the prediction, with higher entropy indicating lower confidence.
% % The loss $\mathcal{L}_\text{Ent}$ optimizes $\tau_\bv^\text{out}$ to reduce the confidence of the specialized prediction $\pred_\bv$ relative to the generalized prediction $\pred_\bh$ for the generated out-classes.
% The loss $\mathcal{L}_\text{Ent}$ optimizes $\tau_\bv^\text{out}$ to reduce the confidence of the specialized prediction $\pred_\bv$ compared with the generalized prediction $\pred_\bh$ by a margin $d$ for the generated out-classes.
% As a result, the weight $\alpha_\text{out}$ for out-class domains is optimized, improving generalization without trade-offs.

\newcommand{\Base}{{\textit{Base}}}
\newcommand{\New}{{\textit{New}}}

%%%%%%%%%%%%%%%%%%%%%%%%%
%%% TABLE 1. Base2New %%%
%%%%%%%%%%%%%%%%%%%%%%%%%

\newcommand{\textBF}[1]{%
    \pdfliteral direct {2 Tr 0.3 w} %the second factor is the boldness
     #1%
    \pdfliteral direct {0 Tr 0 w}%
}

\newcommand{\mc}[1]{\multicolumn{1}{c}{$#1$}}
\newcommand{\bmc}[1]{\multicolumn{1}{c}{$\textBF{#1}$}}
\newcommand{\umc}[1]{\multicolumn{1}{c}{$\underline{#1}$}}
\newcommand{\mcl}[1]{\multicolumn{1}{c|}{$#1$}}
\newcommand{\bmcl}[1]{\multicolumn{1}{c|}{$\textBF{#1}$}}
\newcommand{\umcl}[1]{\multicolumn{1}{c|}{$\underline{#1}$}}
\newcommand{\Datasetl}[1]{\multicolumn{3}{c|}{\textsc{#1}}}
\newcommand{\Dataset}[1]{\multicolumn{3}{c}{\textsc{#1}}}
\newcommand{\Datasetone}[1]{\multicolumn{1}{c}{\textsc{#1}}}
\newcommand{\Datasetonel}[1]{\multicolumn{1}{c|}{\textsc{#1}}}

\newcommand{\MethodFont}[1]{\textsc{#1}}

\vcase{
\begin{table*}[thb!]
    \caption{Performance comparison on 11 datasets in the base-to-new benchmark. 
    H represents the harmonic mean.
    } \label{tab:base2new_generalization}
    \begin{center}
        \resizebox{\linewidth}{!}{
        \begin{threeparttable}
        % \footnotesize{
            \begin{tabular}
                {@{}r | *{3}{c}@{} | *{3}{c}@{} | *{3}{c}@{}}
                \toprule\toprule % \toprule
                 & \Datasetl{Average} & \Datasetl{ImageNet} & \Dataset{Caltech101}
                \\
                % \cmidrule(ll){2-4}\cmidrule(ll){5-7}\cmidrule(ll){8-10}
                \mcl{\MethodFont{Method}}
                & \mc{\MethodFont{Base}} & \mc{\MethodFont{New}} & \mcl{\MethodFont{H}}
                & \mc{\MethodFont{Base}} & \mc{\MethodFont{New}} & \mcl{\MethodFont{H}}
                & \mc{\MethodFont{Base}} & \mc{\MethodFont{New}} & \mc{\MethodFont{H}}
                \\ \midrule
                \mcl{\MethodFont{CLIP}} 
                & \mc{65.14} & \mc{68.78} & \mcl{66.82} % Average
                & \mc{64.43} & \mc{60.04} & \mcl{62.16} % ImageNet
                & \mc{90.64} & \mc{91.16} & \mc{90.90} % Caltech101
                \\
                \mcl{\MethodFont{CoOp}} 
                & \mc{77.23} & \mc{68.56} & \mcl{71.33} % Average
                & \mc{73.72 \pm 0.29} & \mc{64.94 \pm 0.87} & \mcl{69.05} % ImageNet
                & \mc{97.16 \pm 0.16} & \mc{93.92 \pm 0.80} & \mc{95.51} % Caltech101
                \\
                \mcl{\MethodFont{ProGrad}} 
                & \mc{78.74} & \mc{72.19} & \mcl{75.06} % Average
                & \mc{74.81 \pm 0.29} & \mc{66.68 \pm 0.26} & \mcl{70.51} % ImageNet
                & \mc{97.50 \pm 0.08} & \mc{95.49 \pm 0.27} & \mc{96.48} % Caltech101
                \\
                \mcl{\MethodFont{KgCoOp}} 
                & \mc{78.67} & \mc{74.62} & \mcl{76.38} % Average
                % & \mc{75.48 \pm 0.16} & \mc{69.40 \pm 0.20} & \mcl{72.31} % ImageNet
                & \mc{75.44 \pm 0.08} & \mc{69.43 \pm 0.29} & \mcl{72.31} % ImageNet
                & \mc{97.61 \pm 0.33} & \mc{94.80 \pm 0.45} & \mc{96.18} % Caltech101
                \\
                \mcl{\MethodFont{MaPLe}} 
                & \mc{77.14} & \mc{72.91} & \mcl{74.69} % Average
                & \mc{75.40 \pm 0.29} & \bmc{70.43 \pm 0.12} & \bmcl{72.83} % ImageNet
                & \mc{97.47 \pm 0.31} & \mc{93.77 \pm 1.11} & \mc{95.57} % Caltech101
                \\
                \mcl{\MethodFont{DePT}} 
                & \mc{79.20} & \mc{66.36} & \mcl{71.78} % Average
                & \mc{73.50 \pm 0.22} & \mc{70.00 \pm 0.16} & \mcl{71.71} % ImageNet
                & \mc{97.83 \pm 0.05} & \bmc{95.83 \pm 0.25} & \bmc{96.82} % Caltech101
                \\
                \rowcolor{lightgray!50}
                \mcl{\MethodFont{CoA-loss}}  % vit_b16_ep50_v_w5
                & \mc{79.12} & \mc{73.66} & \mcl{76.15} % Average
                & \bmc{75.68 \pm 0.00} & \mc{67.98 \pm 0.31} & \mcl{71.62} % ImageNet
                & \mc{97.94 \pm 0.14} & \mc{94.54 \pm 0.24} & \mc{96.21} % Caltech101
                \\
                \rowcolor{lightgray!50}
                \mcl{\MethodFont{CoCoA-Mix}}  % vit_b16_ep50_v_w5-tin_w1-tout_w10_d02_word
                & \bmc{79.31} & \bmc{75.10} & \bmcl{77.03} % Average
                & \mc{75.47 \pm 0.09} & \mc{68.92 \pm 0.10} & \mcl{72.04} % ImageNet
                & \bmc{98.02 \pm 0.03} & \mc{94.39 \pm 0.10} & \mc{96.17} % Caltech101 
                \\
                % ---------------------------------------------------------------- %
                \toprule\toprule
                 & \Datasetl{OxfordPets} & \Datasetl{StanfordCars} & \Dataset{Flowers102}
                \\
                \mcl{\MethodFont{Method}}
                & \mc{\MethodFont{Base}} & \mc{\MethodFont{New}} & \mcl{\MethodFont{H}}
                & \mc{\MethodFont{Base}} & \mc{\MethodFont{New}} & \mcl{\MethodFont{H}}
                & \mc{\MethodFont{Base}} & \mc{\MethodFont{New}} & \mc{\MethodFont{H}}
                \\ \midrule
                \mcl{\MethodFont{CLIP}} 
                & \mc{90.01} & \mc{94.24} & \mcl{92.07} % OxfordPets
                & \mc{55.37} & \mc{66.65} & \mcl{60.49} % StanfordCars
                & \mc{69.23} & \mc{73.90} & \mc{71.49} % Flowers102
                \\
                \mcl{\MethodFont{CoOp}} 
                & \mc{94.10 \pm 0.73} & \mc{94.42 \pm 4.17} & \mcl{94.16} % OxfordPets
                & \mc{69.54 \pm 0.75} & \mc{71.39 \pm 1.28} & \mcl{70.44} % StanfordCars
                & \mc{90.60 \pm 1.50} & \mc{67.00 \pm 1.04} & \mc{77.01} % Flowers102
                \\
                \mcl{\MethodFont{ProGrad}} 
                & \mc{95.00 \pm 0.31} & \mc{97.36 \pm 0.42} & \mcl{96.16} % OxfordPets
                & \mc{71.45 \pm 0.39} & \mc{73.16 \pm 0.58} & \mcl{72.29} % StanfordCars
                & \mc{91.36 \pm 0.63} & \mc{74.92 \pm 0.90} & \mc{82.32} % Flowers102
                \\
                \mcl{\MethodFont{KgCoOp}} 
                & \mc{94.65 \pm 0.15} & \mc{97.59 \pm 0.08} & \mcl{96.10} % OxfordPets
                & \mc{68.64 \pm 0.35} & \mc{74.96 \pm 0.53} & \mcl{71.66} % StanfordCars
                & \mc{90.09 \pm 0.63} & \mc{76.31 \pm 0.42} & \mc{82.63} % Flowers102
                \\
                \mcl{\MethodFont{MaPLe}} 
                & \mc{94.80 \pm 0.94} & \mc{97.67 \pm 0.21} & \mcl{96.21} % OxfordPets
                & \mc{67.97 \pm 0.29} & \mc{74.40 \pm 0.45} & \mcl{71.04} % StanfordCars
                & \mc{88.03 \pm 1.62} & \mc{73.43 \pm 0.49} & \mc{80.06} % Flowers102
                \\
                \mcl{\MethodFont{DePT}} 
                & \mc{94.00 \pm 0.29} & \mc{88.63 \pm 0.78} & \mcl{91.23} % OxfordPets
                & \mc{71.83 \pm 0.52} & \mc{59.27 \pm 0.76} & \mcl{64.94} % StanfordCars
                & \bmc{94.53 \pm 0.53} & \mc{66.30 \pm 1.42} & \mc{77.92} % Flowers102
                \\
                \rowcolor{lightgray!50}
                \mcl{\MethodFont{CoA-loss}}  % vit_b16_ep50_v_w5
                & \mc{94.90 \pm 0.49} & \bmc{97.93 \pm 0.08} & \bmcl{96.39} % OxfordPets
                & \mc{72.70 \pm 0.11} & \mc{73.07 \pm 1.27} & \mcl{72.87} % StanfordCars
                & \mc{88.89 \pm 1.75} & \mc{75.58 \pm 1.31} & \mc{81.67} % Flowers102
                \\
                \rowcolor{lightgray!50}
                \mcl{\MethodFont{CoCoA-Mix}}  % vit_b16_ep50_v_w5-tin_w1-tout_w10_d02_word
                & \bmc{95.16 \pm 0.38} & \mc{97.60 \pm 0.09} & \mcl{96.36} % OxfordPets
                & \bmc{73.09 \pm 0.25} & \bmc{74.97 \pm 0.08} & \bmcl{74.01} % StanfordCars
                & \mc{91.04 \pm 1.79} & \bmc{77.37 \pm 0.38} & \bmc{83.64} % Flowers102
                \\
                % ---------------------------------------------------------------- %
                \toprule\toprule
                 & \Datasetl{Food101} & \Datasetl{FgvcAircraft} & \Dataset{SUN397}
                \\
                \mcl{\MethodFont{Method}}
                & \mc{\MethodFont{Base}} & \mc{\MethodFont{New}} & \mcl{\MethodFont{H}}
                & \mc{\MethodFont{Base}} & \mc{\MethodFont{New}} & \mcl{\MethodFont{H}}
                & \mc{\MethodFont{Base}} & \mc{\MethodFont{New}} & \mc{\MethodFont{H}}
                \\ \midrule
                \mcl{\MethodFont{CLIP}} 
                & \mc{83.58} & \mc{84.95} & \mcl{84.26} % Food101
                & \mc{19.51} & \mc{24.60} & \mcl{21.76} % FgvcAircraft
                & \mc{66.76} & \mc{70.52} & \mc{68.59} % SUN397
                \\
                \mcl{\MethodFont{CoOp}} 
                & \mc{89.19 \pm 0.19} & \mc{88.45 \pm 0.89} & \mcl{88.81} % Food101
                & \mc{26.17 \pm 7.89} & \mc{19.50 \pm 11.94} & \mcl{11.46} % FgvcAircraft
                & \mc{77.37 \pm 0.66} & \mc{72.06 \pm 1.56} & \mc{74.60} % SUN397
                \\
                \mcl{\MethodFont{ProGrad}} 
                & \mc{89.33 \pm 0.08} & \mc{89.93 \pm 0.58} & \mcl{89.63} % Food101
                & \mc{34.21 \pm 1.99} & \mc{28.53 \pm 2.08} & \mcl{30.97} % FgvcAircraft
                & \bmc{79.16 \pm 0.36} & \mc{74.34 \pm 0.75} & \mc{76.20} % SUN397
                \\
                \mcl{\MethodFont{KgCoOp}} 
                & \bmc{90.26 \pm 0.11} & \bmc{91.25 \pm 0.15} & \bmcl{90.75} % Food101
                & \mc{33.43 \pm 0.56} & \mc{32.27 \pm 1.19} & \mcl{32.81} % FgvcAircraft
                & \mc{79.07 \pm 0.24} & \mc{76.78 \pm 0.24} & \mc{77.91} % SUN397
                \\
                \mcl{\MethodFont{MaPLe}} 
                & \mc{89.37 \pm 0.54} & \mc{90.77 \pm 0.54} & \mcl{90.06} % Food101
                & \mc{31.67 \pm 0.66} & \mc{33.13 \pm 2.38} & \mcl{32.29} % FgvcAircraft
                & \mc{78.33 \pm 0.21} & \bmc{77.67 \pm 0.45} & \bmc{78.00} % SUN397
                \\
                \mcl{\MethodFont{DePT}} 
                & \mc{89.80 \pm 0.08} & \mc{88.10 \pm 0.16} & \mcl{88.94} % Food101
                & \bmc{35.93 \pm 0.93} & \mc{24.33 \pm 0.09} & \mcl{29.01} % FgvcAircraft
                & \mc{79.10 \pm 0.22} & \mc{67.27 \pm 0.46} & \mc{72.70} % SUN397
                \\
                \rowcolor{lightgray!50}
                \mcl{\MethodFont{CoA-loss}}  % vit_b16_ep50_v_w5
                & \mc{90.11 \pm 0.18} & \mc{90.87 \pm 0.42} & \mcl{90.49} % Food101
                & \mc{33.91 \pm 0.68} & \mc{32.47 \pm 0.37} & \mcl{33.17} % FgvcAircraft
                & \mc{78.70 \pm 0.25} & \mc{75.43 \pm 0.72} & \mc{77.03} % SUN397
                \\
                \rowcolor{lightgray!50}
                \mcl{\MethodFont{CoCoA-Mix}}  % vit_b16_ep50_v_w5-tin_w1-tout_w10_d02_word
                & \mc{90.09 \pm 0.16} & \mc{90.93 \pm 0.09} & \mcl{90.50} % Food101
                & \mc{33.51 \pm 0.28} & \bmc{34.15 \pm 0.14} & \bmcl{33.83} % FgvcAircraft
                & \mc{78.51 \pm 0.17} & \mc{76.60 \pm 0.24} & \mc{77.54} % SUN397
                \\
                % ---------------------------------------------------------------- %
                \toprule\toprule
                 & \Datasetl{DTD} & \Datasetl{EuroSAT} & \Dataset{UCF101}
                \\
                \mcl{\MethodFont{Method}}
                & \mc{\MethodFont{Base}} & \mc{\MethodFont{New}} & \mcl{\MethodFont{H}}
                & \mc{\MethodFont{Base}} & \mc{\MethodFont{New}} & \mcl{\MethodFont{H}}
                & \mc{\MethodFont{Base}} & \mc{\MethodFont{New}} & \mc{\MethodFont{H}}
                \\ \midrule
                \mcl{\MethodFont{CLIP}} 
                & \mc{53.24} & \mc{54.71} & \mcl{53.97} % DTD
                & \mc{54.79} & \mc{66.21} & \mcl{59.96} % EuroSAT
                & \mc{69.03} & \mc{69.61} & \mc{69.32} % UCF101
                \\
                \mcl{\MethodFont{CoOp}} 
                & \mc{71.22 \pm 1.13} & \mc{53.62 \pm 3.45} & \mcl{61.03} % DTD
                & \mc{79.93 \pm 1.07} & \mc{64.79 \pm 6.36} & \mcl{71.19} % EuroSAT
                & \mc{80.58 \pm 0.66} & \mc{64.11 \pm 2.84} & \mc{71.32} % UCF101
                \\
                \mcl{\MethodFont{ProGrad}} 
                & \mc{72.07 \pm 0.29} & \mc{50.56 \pm 2.43} & \mcl{59.35} % DTD
                & \mc{81.29 \pm 3.36} & \mc{69.81 \pm 5.56} & \mcl{74.80} % EuroSAT
                & \mc{80.97 \pm 0.29} & \mc{73.32 \pm 1.85} & \mc{76.93} % UCF101
                \\
                \mcl{\MethodFont{KgCoOp}} 
                & \mc{72.92 \pm 1.05} & \mc{59.14 \pm 1.53} & \mcl{65.28} % DTD
                & \mc{83.20 \pm 0.72} & \bmc{70.51 \pm 9.30} & \bmcl{75.61} % EuroSAT
                & \mc{80.09 \pm 0.24} & \bmc{77.75 \pm 0.40} & \mc{78.90} % UCF101
                \\
                \mcl{\MethodFont{MaPLe}} 
                & \mc{70.40 \pm 2.57} & \mc{58.40 \pm 3.00} & \mcl{63.71} % DTD
                & \mc{76.50 \pm 3.85} & \mc{55.70 \pm 3.19} & \mcl{64.27} % EuroSAT
                & \mc{78.57 \pm 2.11} & \mc{76.60 \pm 1.56} & \mc{77.53} % UCF101
                \\
                \mcl{\MethodFont{DePT}} 
                & \bmc{74.40 \pm 0.83} & \mc{53.13 \pm 1.07} & \mcl{61.98} % DTD
                & \mc{78.70 \pm 1.56} & \mc{50.53 \pm 5.71} & \mcl{61.08} % EuroSAT
                & \bmc{81.57 \pm 0.84} & \mc{66.53 \pm 0.87} & \mc{73.28} % UCF101
                \\
                \rowcolor{lightgray!50}
                \mcl{\MethodFont{CoA-loss}}  % vit_b16_ep50_v_w5
                & \mc{73.23 \pm 2.02} & \mc{58.09 \pm 0.81} & \mcl{64.76} % DTD
                & \mc{83.38 \pm 0.49} & \mc{70.07 \pm 2.49} & \mcl{76.09} % EuroSAT
                & \mc{80.83 \pm 0.80} & \mc{74.22 \pm 0.91} & \mc{77.38} % UCF101
                \\
                \rowcolor{lightgray!50}
                \mcl{\MethodFont{CoCoA-Mix}}  % vit_b16_ep50_v_w5-tin_w1-tout_w10_d02_word
                & \mc{72.80 \pm 1.89} & \bmc{64.29 \pm 1.25} & \bmcl{68.25} % DTD
                & \bmc{83.49 \pm 0.66} & \mc{69.11 \pm 3.10} & \mcl{75.54} % EuroSAT
                & \mc{81.28 \pm 0.95} & \bmc{77.75 \pm 0.24} & \bmc{79.47} % UCF101
                \\
                \bottomrule \bottomrule
            \end{tabular}
        % }
        \end{threeparttable}
        }
    \end{center}
\end{table*}
}{
\begin{table*}[thb!]
    \caption{Performance comparison on 11 datasets in the base-to-new benchmark. 
    H represents the harmonic mean.
    } \label{tab:base2new_generalization}
    \begin{center}
        \resizebox{\linewidth}{!}{
        \begin{threeparttable}
        % \footnotesize{
            \begin{tabular}
                {@{}r | *{3}{c}@{} | *{3}{c}@{} | *{3}{c}@{}}
                \toprule\toprule % \toprule
                 & \Datasetl{Average} & \Datasetl{ImageNet} & \Dataset{Caltech101}
                \\
                % \cmidrule(ll){2-4}\cmidrule(ll){5-7}\cmidrule(ll){8-10}
                \mcl{\MethodFont{Method}}
                & \mc{\MethodFont{Base}} & \mc{\MethodFont{New}} & \mcl{\MethodFont{H}}
                & \mc{\MethodFont{Base}} & \mc{\MethodFont{New}} & \mcl{\MethodFont{H}}
                & \mc{\MethodFont{Base}} & \mc{\MethodFont{New}} & \mc{\MethodFont{H}}
                \\ \midrule
                \mcl{\MethodFont{CLIP}} 
                & \mc{65.14} & \mc{68.78} & \mcl{66.82} % Average
                & \mc{64.43} & \mc{60.04} & \mcl{62.16} % ImageNet
                & \mc{90.64} & \mc{91.16} & \mc{90.90} % Caltech101
                \\
                \mcl{\MethodFont{CoOp}} 
                & \mc{77.23} & \mc{68.56} & \mcl{71.33} % Average
                & \mc{73.72 \pm 0.29} & \mc{64.94 \pm 0.87} & \mcl{69.05} % ImageNet
                & \mc{97.16 \pm 0.16} & \mc{93.92 \pm 0.80} & \mc{95.51} % Caltech101
                \\
                \mcl{\MethodFont{ProGrad}} 
                & \mc{78.74} & \mc{72.19} & \mcl{75.06} % Average
                & \mc{74.81 \pm 0.29} & \mc{66.68 \pm 0.26} & \mcl{70.51} % ImageNet
                & \mc{97.50 \pm 0.08} & \mc{95.49 \pm 0.27} & \mc{96.48} % Caltech101
                \\
                \mcl{\MethodFont{KgCoOp}} 
                & \mc{78.67} & \mc{74.62} & \mcl{76.38} % Average
                % & \mc{75.48 \pm 0.16} & \mc{69.40 \pm 0.20} & \mcl{72.31} % ImageNet
                & \mc{75.44 \pm 0.08} & \mc{69.43 \pm 0.29} & \mcl{72.31} % ImageNet
                & \mc{97.61 \pm 0.33} & \mc{94.80 \pm 0.45} & \mc{96.18} % Caltech101
                \\
                \mcl{\MethodFont{MaPLe}} 
                & \mc{77.14} & \mc{72.91} & \mcl{74.69} % Average
                & \mc{75.40 \pm 0.29} & \bmc{70.43 \pm 0.12} & \bmcl{72.83} % ImageNet
                & \mc{97.47 \pm 0.31} & \mc{93.77 \pm 1.11} & \mc{95.57} % Caltech101
                \\
                \mcl{\MethodFont{DePT}} 
                & \mc{79.20} & \mc{66.36} & \mcl{71.78} % Average
                & \mc{73.50 \pm 0.22} & \mc{70.00 \pm 0.16} & \mcl{71.71} % ImageNet
                & \mc{97.83 \pm 0.05} & \bmc{95.83 \pm 0.25} & \bmc{96.82} % Caltech101
                \\
                \rowcolor{lightgray!50}
                \mcl{\MethodFont{CoA-loss}}  % vit_b16_ep50_v_w5
                & \mc{79.12} & \mc{73.66} & \mcl{76.15} % Average
                & \bmc{75.68 \pm 0.00} & \mc{67.98 \pm 0.31} & \mcl{71.62} % ImageNet
                & \mc{97.94 \pm 0.14} & \mc{94.54 \pm 0.24} & \mc{96.21} % Caltech101
                \\
                \rowcolor{lightgray!50}
                \mcl{\MethodFont{CoCoA-Mix}}  % 
                & \bmc{79.35} & \bmc{75.04} & \bmcl{76.98} % Average
                & \mc{74.15 \pm 0.20} & \mc{68.52 \pm 0.05} & \mcl{71.22} % ImageNet
                & \bmc{97.93 \pm 0.14} & \mc{94.43 \pm 0.09} & \mc{96.15} % Caltech101 
                \\
                % ---------------------------------------------------------------- %
                \toprule\toprule
                 & \Datasetl{OxfordPets} & \Datasetl{StanfordCars} & \Dataset{Flowers102}
                \\
                \mcl{\MethodFont{Method}}
                & \mc{\MethodFont{Base}} & \mc{\MethodFont{New}} & \mcl{\MethodFont{H}}
                & \mc{\MethodFont{Base}} & \mc{\MethodFont{New}} & \mcl{\MethodFont{H}}
                & \mc{\MethodFont{Base}} & \mc{\MethodFont{New}} & \mc{\MethodFont{H}}
                \\ \midrule
                \mcl{\MethodFont{CLIP}} 
                & \mc{90.01} & \mc{94.24} & \mcl{92.07} % OxfordPets
                & \mc{55.37} & \mc{66.65} & \mcl{60.49} % StanfordCars
                & \mc{69.23} & \mc{73.90} & \mc{71.49} % Flowers102
                \\
                \mcl{\MethodFont{CoOp}} 
                & \mc{94.10 \pm 0.73} & \mc{94.42 \pm 4.17} & \mcl{94.16} % OxfordPets
                & \mc{69.54 \pm 0.75} & \mc{71.39 \pm 1.28} & \mcl{70.44} % StanfordCars
                & \mc{90.60 \pm 1.50} & \mc{67.00 \pm 1.04} & \mc{77.01} % Flowers102
                \\
                \mcl{\MethodFont{ProGrad}} 
                & \mc{95.00 \pm 0.31} & \mc{97.36 \pm 0.42} & \mcl{96.16} % OxfordPets
                & \mc{71.45 \pm 0.39} & \mc{73.16 \pm 0.58} & \mcl{72.29} % StanfordCars
                & \mc{91.36 \pm 0.63} & \mc{74.92 \pm 0.90} & \mc{82.32} % Flowers102
                \\
                \mcl{\MethodFont{KgCoOp}} 
                & \mc{94.65 \pm 0.15} & \mc{97.59 \pm 0.08} & \mcl{96.10} % OxfordPets
                & \mc{68.64 \pm 0.35} & \mc{74.96 \pm 0.53} & \mcl{71.66} % StanfordCars
                & \mc{90.09 \pm 0.63} & \mc{76.31 \pm 0.42} & \mc{82.63} % Flowers102
                \\
                \mcl{\MethodFont{MaPLe}} 
                & \mc{94.80 \pm 0.94} & \mc{97.67 \pm 0.21} & \mcl{96.21} % OxfordPets
                & \mc{67.97 \pm 0.29} & \mc{74.40 \pm 0.45} & \mcl{71.04} % StanfordCars
                & \mc{88.03 \pm 1.62} & \mc{73.43 \pm 0.49} & \mc{80.06} % Flowers102
                \\
                \mcl{\MethodFont{DePT}} 
                & \mc{94.00 \pm 0.29} & \mc{88.63 \pm 0.78} & \mcl{91.23} % OxfordPets
                & \mc{71.83 \pm 0.52} & \mc{59.27 \pm 0.76} & \mcl{64.94} % StanfordCars
                & \bmc{94.53 \pm 0.53} & \mc{66.30 \pm 1.42} & \mc{77.92} % Flowers102
                \\
                \rowcolor{lightgray!50}
                \mcl{\MethodFont{CoA-loss}}  % vit_b16_ep50_v_w5
                & \mc{94.90 \pm 0.49} & \bmc{97.93 \pm 0.08} & \bmcl{96.39} % OxfordPets
                & \mc{72.70 \pm 0.11} & \mc{73.07 \pm 1.27} & \mcl{72.87} % StanfordCars
                & \mc{88.89 \pm 1.75} & \mc{75.58 \pm 1.31} & \mc{81.67} % Flowers102
                \\
                \rowcolor{lightgray!50}
                \mcl{\MethodFont{CoCoA-Mix}} 
                & \bmc{95.14 \pm 0.41} & \mc{97.45 \pm 0.11} & \mcl{96.28} % OxfordPets
                & \bmc{72.91 \pm 0.19} & \bmc{75.18 \pm 0.21} & \bmcl{74.03} % StanfordCars
                & \mc{92.37 \pm 1.71} & \bmc{77.40 \pm 0.13} & \bmc{84.21} % Flowers102
                \\
                % ---------------------------------------------------------------- %
                \toprule\toprule
                 & \Datasetl{Food101} & \Datasetl{FgvcAircraft} & \Dataset{SUN397}
                \\
                \mcl{\MethodFont{Method}}
                & \mc{\MethodFont{Base}} & \mc{\MethodFont{New}} & \mcl{\MethodFont{H}}
                & \mc{\MethodFont{Base}} & \mc{\MethodFont{New}} & \mcl{\MethodFont{H}}
                & \mc{\MethodFont{Base}} & \mc{\MethodFont{New}} & \mc{\MethodFont{H}}
                \\ \midrule
                \mcl{\MethodFont{CLIP}} 
                & \mc{83.58} & \mc{84.95} & \mcl{84.26} % Food101
                & \mc{19.51} & \mc{24.60} & \mcl{21.76} % FgvcAircraft
                & \mc{66.76} & \mc{70.52} & \mc{68.59} % SUN397
                \\
                \mcl{\MethodFont{CoOp}} 
                & \mc{89.19 \pm 0.19} & \mc{88.45 \pm 0.89} & \mcl{88.81} % Food101
                & \mc{26.17 \pm 7.89} & \mc{19.50 \pm 11.94} & \mcl{11.46} % FgvcAircraft
                & \mc{77.37 \pm 0.66} & \mc{72.06 \pm 1.56} & \mc{74.60} % SUN397
                \\
                \mcl{\MethodFont{ProGrad}} 
                & \mc{89.33 \pm 0.08} & \mc{89.93 \pm 0.58} & \mcl{89.63} % Food101
                & \mc{34.21 \pm 1.99} & \mc{28.53 \pm 2.08} & \mcl{30.97} % FgvcAircraft
                & \bmc{79.16 \pm 0.36} & \mc{74.34 \pm 0.75} & \mc{76.20} % SUN397
                \\
                \mcl{\MethodFont{KgCoOp}} 
                & \bmc{90.26 \pm 0.11} & \bmc{91.25 \pm 0.15} & \bmcl{90.75} % Food101
                & \mc{33.43 \pm 0.56} & \mc{32.27 \pm 1.19} & \mcl{32.81} % FgvcAircraft
                & \mc{79.07 \pm 0.24} & \mc{76.78 \pm 0.24} & \mc{77.91} % SUN397
                \\
                \mcl{\MethodFont{MaPLe}} 
                & \mc{89.37 \pm 0.54} & \mc{90.77 \pm 0.54} & \mcl{90.06} % Food101
                & \mc{31.67 \pm 0.66} & \mc{33.13 \pm 2.38} & \mcl{32.29} % FgvcAircraft
                & \mc{78.33 \pm 0.21} & \bmc{77.67 \pm 0.45} & \bmc{78.00} % SUN397
                \\
                \mcl{\MethodFont{DePT}} 
                & \mc{89.80 \pm 0.08} & \mc{88.10 \pm 0.16} & \mcl{88.94} % Food101
                & \bmc{35.93 \pm 0.93} & \mc{24.33 \pm 0.09} & \mcl{29.01} % FgvcAircraft
                & \mc{79.10 \pm 0.22} & \mc{67.27 \pm 0.46} & \mc{72.70} % SUN397
                \\
                \rowcolor{lightgray!50}
                \mcl{\MethodFont{CoA-loss}}  % vit_b16_ep50_v_w5
                & \mc{90.11 \pm 0.18} & \mc{90.87 \pm 0.42} & \mcl{90.49} % Food101
                & \mc{33.91 \pm 0.68} & \mc{32.47 \pm 0.37} & \mcl{33.17} % FgvcAircraft
                & \mc{78.70 \pm 0.25} & \mc{75.43 \pm 0.72} & \mc{77.03} % SUN397
                \\
                \rowcolor{lightgray!50}
                \mcl{\MethodFont{CoCoA-Mix}} 
                & \mc{89.87 \pm 0.06} & \mc{90.85 \pm 0.07} & \mcl{90.36} % Food101
                & \mc{33.19 \pm 0.76} & \bmc{34.45 \pm 0.48} & \bmcl{33.80} % FgvcAircraft
                & \mc{78.27 \pm 0.31} & \mc{76.20 \pm 0.12} & \mc{77.22} % SUN397
                \\
                % ---------------------------------------------------------------- %
                \toprule\toprule
                 & \Datasetl{DTD} & \Datasetl{EuroSAT} & \Dataset{UCF101}
                \\
                \mcl{\MethodFont{Method}}
                & \mc{\MethodFont{Base}} & \mc{\MethodFont{New}} & \mcl{\MethodFont{H}}
                & \mc{\MethodFont{Base}} & \mc{\MethodFont{New}} & \mcl{\MethodFont{H}}
                & \mc{\MethodFont{Base}} & \mc{\MethodFont{New}} & \mc{\MethodFont{H}}
                \\ \midrule
                \mcl{\MethodFont{CLIP}} 
                & \mc{53.24} & \mc{54.71} & \mcl{53.97} % DTD
                & \mc{54.79} & \mc{66.21} & \mcl{59.96} % EuroSAT
                & \mc{69.03} & \mc{69.61} & \mc{69.32} % UCF101
                \\
                \mcl{\MethodFont{CoOp}} 
                & \mc{71.22 \pm 1.13} & \mc{53.62 \pm 3.45} & \mcl{61.03} % DTD
                & \mc{79.93 \pm 1.07} & \mc{64.79 \pm 6.36} & \mcl{71.19} % EuroSAT
                & \mc{80.58 \pm 0.66} & \mc{64.11 \pm 2.84} & \mc{71.32} % UCF101
                \\
                \mcl{\MethodFont{ProGrad}} 
                & \mc{72.07 \pm 0.29} & \mc{50.56 \pm 2.43} & \mcl{59.35} % DTD
                & \mc{81.29 \pm 3.36} & \mc{69.81 \pm 5.56} & \mcl{74.80} % EuroSAT
                & \mc{80.97 \pm 0.29} & \mc{73.32 \pm 1.85} & \mc{76.93} % UCF101
                \\
                \mcl{\MethodFont{KgCoOp}} 
                & \mc{72.92 \pm 1.05} & \mc{59.14 \pm 1.53} & \mcl{65.28} % DTD
                & \mc{83.20 \pm 0.72} & \bmc{70.51 \pm 9.30} & \bmcl{75.61} % EuroSAT
                & \mc{80.09 \pm 0.24} & \bmc{77.75 \pm 0.40} & \mc{78.90} % UCF101
                \\
                \mcl{\MethodFont{MaPLe}} 
                & \mc{70.40 \pm 2.57} & \mc{58.40 \pm 3.00} & \mcl{63.71} % DTD
                & \mc{76.50 \pm 3.85} & \mc{55.70 \pm 3.19} & \mcl{64.27} % EuroSAT
                & \mc{78.57 \pm 2.11} & \mc{76.60 \pm 1.56} & \mc{77.53} % UCF101
                \\
                \mcl{\MethodFont{DePT}} 
                & \bmc{74.40 \pm 0.83} & \mc{53.13 \pm 1.07} & \mcl{61.98} % DTD
                & \mc{78.70 \pm 1.56} & \mc{50.53 \pm 5.71} & \mcl{61.08} % EuroSAT
                & \bmc{81.57 \pm 0.84} & \mc{66.53 \pm 0.87} & \mc{73.28} % UCF101
                \\
                \rowcolor{lightgray!50}
                \mcl{\MethodFont{CoA-loss}}  % vit_b16_ep50_v_w5
                & \mc{73.23 \pm 2.02} & \mc{58.09 \pm 0.81} & \mcl{64.76} % DTD
                & \mc{83.38 \pm 0.49} & \mc{70.07 \pm 2.49} & \mcl{76.09} % EuroSAT
                & \mc{80.83 \pm 0.80} & \mc{74.22 \pm 0.91} & \mc{77.38} % UCF101
                \\
                \rowcolor{lightgray!50}
                \mcl{\MethodFont{CoCoA-Mix}}
                & \mc{73.50 \pm 1.80} & \bmc{64.86 \pm 1.04} & \bmcl{68.88} % DTD
                & \bmc{84.20 \pm 0.65} & \mc{67.58 \pm 4.86} & \mcl{74.76} % EuroSAT
                & \mc{81.28 \pm 0.55} & \bmc{78.53 \pm 0.31} & \bmc{79.88} % UCF101
                \\
                \bottomrule \bottomrule
            \end{tabular}
        % }
        \end{threeparttable}
        }
    \end{center}
\end{table*}
}

\section{Experiments}\label{sec:experiments}

%%%%%%%%%%%%%%%%%%%%%%%%%%%%%%%%%
%%% Version 1.4. (2025.01.01) %%%
%%%%%%%%%%%%%%%%%%%%%%%%%%%%%%%%%

%-------------------------------------------------------------------------
\subsection{Datasets and Implementation Details}

% We validate the effectiveness of CoCoA-Mix in three benchmarks: 
We validate the effectiveness of our method in three tasks: 
(1) base-to-new generalization, (2) few-shot class-incremental learning (FSCIL), and (3) cross-dataset transfer. 
%(1) few-shot classification, (2) domain generalization, (3) base-to-new generalization, (4) cross-dataset transfer, and (5) class-incremental learning. 

\paragraph{Datasets}

% We evaluate base-to-new generalization and cross-dataset transfer performance using 11 diverse benchmark datasets: ImageNet~\cite{deng2009imagenet}, Caltech101~\cite{fei2004learning}, OxfordPets~\cite{parkhi2012cats}, StanfordCars~\cite{krause20133d}, Flowers102~\cite{nilsback2008automated}, Food101~\cite{bossard2014food}, FGVCAircraft~\cite{maji2013fine}, EuroSAT~\cite{helber2019eurosat}, UCF101~\cite{soomro2012ucf101}, DTD~\cite{cimpoi2014describing}, and SUN397~\cite{xiao2010sun}. 
We evaluate base-to-new generalization and cross-dataset transfer performance using 11 datasets: ImageNet~\cite{deng2009imagenet}, Caltech101~\cite{fei2004learning}, OxfordPets~\cite{parkhi2012cats}, StanfordCars~\cite{krause20133d}, Flowers102~\cite{nilsback2008automated}, Food101~\cite{bossard2014food}, FGVCAircraft~\cite{maji2013fine}, EuroSAT~\cite{helber2019eurosat}, UCF101~\cite{soomro2012ucf101}, DTD~\cite{cimpoi2014describing}, and SUN397~\cite{xiao2010sun}. 
% These datasets cover various tasks, including generic object classification, fine-grained recognition, texture classification, and scene understanding.
% 자세한 dataset 분류는 appendix에 있다.
% For domain generalization, we use ImageNet as the source dataset and evaluate on ImageNet-V2~\cite{recht2019imagenet}, ImageNet-Sketch~\cite{wang2019learning}, ImageNet-A~\cite{hendrycks2021natural}, and ImageNet-R~\cite{hendrycks2021many} as target datasets. These datasets test the robustness of our method under varying domain shifts.~\todo{transfer and CIL}
% For few-shot class-incremental learning (FSCIL), we use standardized benchmark datasets: CIFAR100~\cite{krizhevsky2009learning}. % , miniImageNet~\cite{krizhevsky2012imagenet}, and CUB200~\cite{wah2011caltech}. 
% Each dataset is split into standard task increments, evaluating the ability of the model to retain knowledge of previously trained classes while adapting to newly introduced ones.
For FSCIL, we use CIFAR100~\cite{krizhevsky2009learning}.
Following Tao et al.~\yrcite{tao2020few}, we split the classes into 60 {\Base} and 40 {\New} classes and adopted a 5-shot 5-way setting, resulting in a total of 9 training sessions.

\paragraph{Training Details}
The prompt length $M$ is initialized randomly and set to $16$ unless specified. % The temperature parameters are initialized to the temperature value from the pre-trained CLIP. 
The out-class set $\mathcal{Y}_i^\text{out}$ for optimizing \vtwo{$\pi_i^\text{out}$} is generated by sampling the same number of random words as the in-class set $\mathcal{Y}_i$ using the API~\cite{wonderwords}. % ~\cite{randomwordapi}.
Prompt tuning is performed using the Adam optimizer~\cite{kingma2014adam} with a learning rate of $0.002$. % and follows a cosine decay schedule. 
Optimization for the \vtwo{CoA-weights} is conducted with SGD. 
% The batch size is $32$ and training is performed over $50$ epochs.
% For a fair evaluation, the final performance is reported as the average of three random seeds ($1, 2, 3$).
Further details of implementation are provided in~\cref{sec:appendix-implementation_details}.

%-------------------------------------------------------------------------
\subsection{Performance Comparison}

\paragraph{Base-to-New Generalization}

% We evaluate the specialization and generalization performance of prompt tuning over classes. 
We evaluate prompt tuning performance over classes in a 4-shot setting.
Each dataset is evenly split into two disjoint subsets: base classes ({\Base}) for tuning and unseen new classes ({\New}). 
Accuracy is measured independently on {\Base} and {\New}, and their harmonic mean H~\cite{xian2017zero} is calculated to evaluate the trade-off between them.
The final performance is reported as the average of three random seeds for a fair evaluation. % ($1, 2, 3$).

\tabref{tab:base2new_generalization} shows that CoOp~\cite{zhou2022learning} achieves a higher average performance than CLIP on {\Base} but a lower average performance on {\New}, highlighting the need for generalization in prompt tuning.
ProGrad~\cite{zhu2023prompt} and KgCoOp~\cite{yao2023visual}, which incorporate hand-crafted prompts during training, improve average performance on both {\Base} and {\New}. 
% However, KgCoOP, which regularizes the prompt tuning, shows reduced specialization on datasets like \texttt{StanfordCars} and \texttt{Flowers102}, suggesting that the prior knowledge of hand-crafted prompts can limit specialization.
% TCP$^\dagger$ refers to the original version, while TCP$^\ddagger$ represents a version with hyperparameters optimized for the current setting.
% TCP~\cite{yao2024tcp} suffers from reduced generalization due to overfitting caused by additional parameters when only 4-shot training samples are available.
MaPLe~\cite{khattak2023maple} and DePT~\cite{zhang2024dept} focus on improving performance on both {\Base} and {\New} by increasing model capacity.
However, with only $4$-shot training samples, they often overfit, limiting generalization.
% Our CoA-loss achieves \TBU{the highest average performance} on {\Base} among existing methods but shows limited generalization as it only focuses on specialization.
% \TBU{Our CoA-loss enhances the {\Base} performance effectively but shows limited generalization as it only focuses on specialization.}
Our CoA-loss, when combined with a naive ensemble, improves {\Base} performance but offers limited generalization as it only focuses on specialization.
By incorporating \vtwo{CoA-weights}, our CoCoA-Mix achieves the highest average performance on both {\Base} and {\New} without trade-offs. 
Notably, CoCoA-Mix surpasses the specialization performance of DePT while using $2.8\%$ of its parameters.
% The proposed CoCoA-Mix performs the best across most datasets, effectively improving specialization and generalization without trade-offs.
% Paired t-test results with a p-value of \TBU{0.xxx} confirm the statistical significance of performance improvements across datasets.

%%%%%%%%%%%%%%%%%%%%%%
%%% TABLE 3. FSCIL %%%
%%%%%%%%%%%%%%%%%%%%%%

\begin{table*}[thb!]
    \begin{center}
        \caption{
        Performance comparison on CIFAR100 in the FSCIL benchmark.
        Mean represents the average accuracy across all sessions, and PD indicates the performance difference between the first and last sessions.
        } \label{tab:fscil}
        \begin{threeparttable}
        \footnotesize{
            \begin{tabular}
                {@{}r | *{9}{c}@{} | *{2}{c}@{}}
                \toprule\toprule % \toprule
                % \multicolumn{12}{c@{}}{\textsc{CIFAR100}}\\
                % \toprule
                \multicolumn{1}{c|@{}}{\multirow{2}{*}{\MethodFont{Method}}}
                & \multicolumn{9}{c|@{}}{\MethodFont{ACC(\%)↑}} 
                & \multicolumn{1}{c@{}}{\multirow{2}{*}{\MethodFont{Mean↑}}} 
                & \multicolumn{1}{c@{}}{\multirow{2}{*}{\MethodFont{PD↓}}} 
                \\
                % \cmidrule(ll){2-4}\cmidrule(ll){5-7}\cmidrule(ll){8-10}
                & \mc{0} & \mc{1} & \mc{2} & \mc{3} & \mc{4}
                & \mc{5} & \mc{6} & \mc{7} & \mcl{8}
                &        &
                \\ \midrule
                \mcl{\MethodFont{L2P}} 
                & \bmc{89.9} & \bmc{86.0} & \mc{81.8} % 0 1 2
                & \mc{80.3} & \mc{80.0} & \mc{74.6} % 3 4 5
                & \mc{73.2} & \mc{72.6} & \mcl{65.0} % 6 7 8
                & \mc{78.2} & \mc{24.9} % Mean      % PD
                \\
                \mcl{\MethodFont{CLIP-ZSL}} 
                & \mc{-} & \mc{-} & \mc{-} % 0 1 2
                & \mc{-} & \mc{-} & \mc{-} % 3 4 5
                & \mc{-} & \mc{-} & \mcl{-} % 6 7 8
                & \mc{77.9} & \mc{-} % Mean      % PD
                \\
                \mcl{\MethodFont{CoOp-FSCIL}} 
                & \mc{88.6} & \mc{78.9} & \mc{77.5} % 0 1 2
                & \mc{76.0} & \mc{76.8} & \mc{78.3} % 3 4 5
                & \mc{79.2} & \mc{79.8} & \mcl{79.3} % 6 7 8
                & \mc{79.4} & \mc{9.3} % Mean      % PD
                \\
                \mcl{\MethodFont{FACT w/ CLIP}} 
                & \mc{87.8} & \mc{84.0} & \mc{81.4} % 0 1 2
                & \mc{78.0} & \mc{77.8} & \mc{76.3} % 3 4 5
                & \mc{75.0} & \mc{72.5} & \mcl{71.9} % 6 7 8
                & \mc{78.3} & \mc{15.9} % Mean      % PD
                \\
                \mcl{\MethodFont{FSPT-FSCIL}} 
                & \mc{86.9} & \mc{83.1} & \mc{81.9} % 0 1 2
                & \mc{80.7} & \mc{80.4} & \mc{79.9} % 3 4 5
                & \mc{80.1} & \mc{79.9} & \mcl{79.4} % 6 7 8
                & \mc{81.4} & \mc{7.5} % Mean      % PD
                \\
                %%% v1
                % \rowcolor{lightgray!50}
                % \mcl{\MethodFont{CoCoA-Mix (Ours)}}  % vit_l14_ep50_v_w5-tin_w1-tout_w10_d02_word_fscil (epoch=10&5, lr=0.002)
                % & \mc{86.9} & \mc{84.5} & \bmc{83.8} % 0 1 2
                % & \bmc{81.7} & \bmc{81.3} & \bmc{81.0} % 3 4 5
                % & \bmc{81.1} & \bmc{80.7} & \bmcl{79.8} % 6 7 8
                % & \bmc{82.3} & \bmc{7.1} % Mean      % PD
                % \\
                %%% v2
                \rowcolor{lightgray!50}
                \mcl{\MethodFont{CoCoA-Mix (Ours)}}
                % CoCoA-Mix (vit_l14_ep50--nctx2) 
                % max_ep=2, max_tune_ep=100
                % i1.0-o10.0-d0.1-e50_2_100-lr002
                & \mc{88.2} & \mc{85.6} & \bmc{84.6} % 0 1 2
                & \bmc{82.7} & \bmc{82.8} & \bmc{82.5} % 3 4 5
                & \bmc{82.3} & \bmc{81.8} & \bmcl{80.8} % 6 7 8
                & \bmc{83.5} & \bmc{7.4} % Mean      % PD
                \\
                \bottomrule \bottomrule
            \end{tabular}
        }
        \end{threeparttable}
    \end{center}
\end{table*}

\paragraph{Few-Shot Class-Incremental Learning (FSCIL)}

We evaluated our method on FSCIL tasks to verify its effectiveness in incremental learning with limited data.
In FSCIL, the number of prompts $K$ was increased incrementally, with each prompt specializing in its session.
For a fair comparison, CoCoA-Mix uses prompts with $M=2$ per session. It has fewer parameters than methods with $M=16$ until session 6 but requires more parameters from session 7.
\cref{tab:fscil} shows the performance of FSCIL on the CIFAR100 dataset.
L2P~\cite{wang2022learning}, which dynamically selects prompts from a pool, shows high initial performance but suffers significant knowledge forgetting as new classes are added, leading to the highest performance difference (PD).
CoOp-FSCIL~\cite{zhou2022learning} and FACT w/ CLIP~\cite{zhou2022forward} outperform zero-shot CLIP in early sessions but are affected by knowledge forgetting in later sessions.
\EDITED{
FSPT-FSCIL~\cite{ran2024brain} outperforms zero-shot CLIP in all sessions by leveraging a brain-inspired strategy in prompt tuning.
}
% In contrast, CoCoA-Mix achieves the highest performance across most sessions, outperforming the state-of-the-art methods.
CoCoA-Mix performs lower in the first two sessions due to fewer parameters but achieves the highest performance in later sessions, outperforming the state-of-the-art methods.
% Existing methods~\cite{wang2022learning, zhou2022learning, zhou2022forward, ran2024brain} often suffer from performance degradation when new classes are incrementally added, because they fail to retain specialization on earlier tasks and generalize to new tasks at the same time.
% In contrast, our CoCoA-Mix leverages CoA-loss to optimize prompts specialized for the given class and applies CoA-temp to scale predictions based on class-specific confidence.
% This approach achieves the highest mean and PD performance without forgetting. 
% As a result, the proposed CoCoA-Mix demonstrates its effectiveness for FSCIL scenarios.
Details of the FSCIL implementation are provided in~\cref{sec:implementation_details-fscil}.

\paragraph{Cross-Dataset Transfer}

%%%%%%%%%%%%%%%%%%%%%%%%%%%%%%
%%% Cross-Dataset Transfer %%%
%%%%%%%%%%%%%%%%%%%%%%%%%%%%%%
\begin{table}[tb]
    \caption{Performance comparison in cross-dataset transfer} \label{tab:transfer}
    \begin{center}
        \begin{threeparttable}
        \footnotesize{
        \begin{tabular}
            % {@{}r | *{1}{c}@{} *{1}{c}@{} *{1}{c}@{}}
            {@{}r | c c c@{}}
            \toprule%\toprule
             \mcl{\MethodFont{Method}}
             & \mc{\MethodFont{Source}} 
             & \mc{\MethodFont{Target}}
             & \mc{\MethodFont{H}}
             \\
            \midrule
            % --------------------------------------- %
            \mcl{\MethodFont{CLIP}} & \mc{66.73 } & \mc{64.89} & \mc{63.97}
            \\
            % --------------------------------------- %
            \mcl{\MethodFont{CoOp}} & \mc{69.06 \pm 0.43} & \mc{59.88} & \mc{61.52}
            \\
            % --------------------------------------- %
            \mcl{\MethodFont{ProGrad}} & \mc{70.21 \pm 0.16} & \mc{62.36} & \mc{63.58}
            \\
            % --------------------------------------- %
            \mcl{\MethodFont{KgCoOp}} & \mc{70.52 \pm 0.05} & \mc{64.45} & \mc{65.17}
            \\
            % --------------------------------------- %
            \mcl{\MethodFont{MaPLe}} & \mc{69.53 \pm 0.39} & \mc{65.24} & \mc{65.26}
            \\
            % --------------------------------------- %
            \mcl{\MethodFont{DePT}} & \mc{68.03 \pm 0.09} & \mc{65.06} & \mc{64.42}
            \\
            % --------------------------------------- %
            \rowcolor{lightgray!50}
            \mcl{\MethodFont{CoCoA-Mix}} & \bmc{70.85 \pm 0.09} & \bmc{65.27} & \bmc{66.07}
            \\
            \bottomrule%\bottomrule
        \end{tabular}
        }        
        \end{threeparttable}
    \end{center}
\end{table}

\begin{figure}[tb!]
    \centering
    \includegraphics[width=0.40\textwidth]{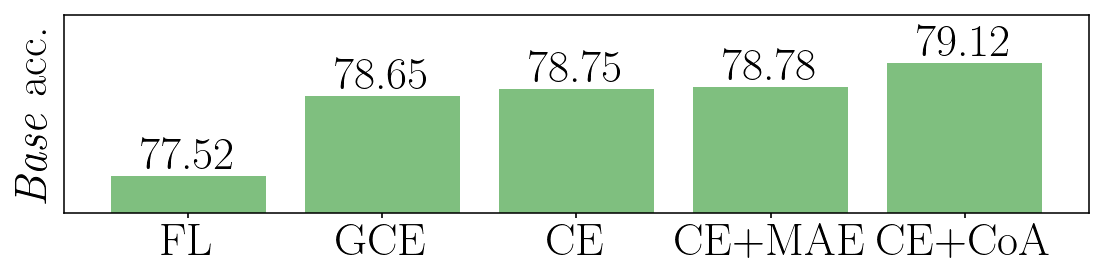}
    \caption{
    Comparison of {\Base} performance by loss function
    }
    \label{fig:lossfunctions}
\end{figure}

\begin{figure*}[!t]
    \centering  
    % \subfigure[]{\label{fig:coaloss_gradnorm}\includegraphics[width=40mm]{resources/CoAloss_gradnorm.png}}
    \subfigure[]{\label{fig:coaloss_ratio}\includegraphics[height=36mm]{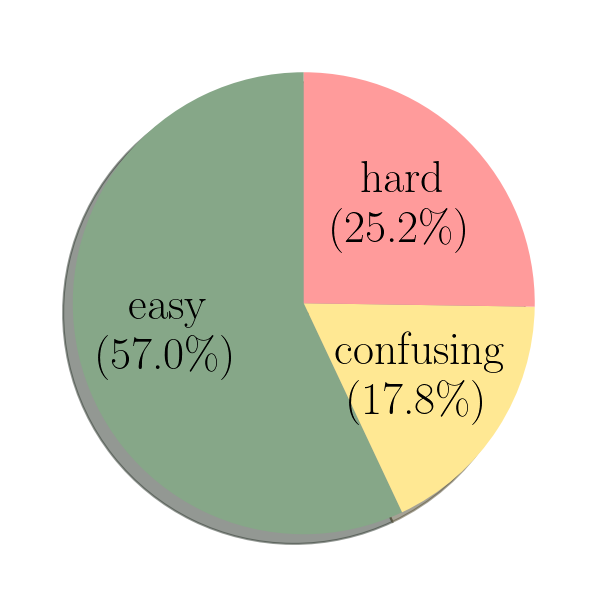}}
    \subfigure[]{\label{fig:coaloss_correct}\includegraphics[width=40mm]{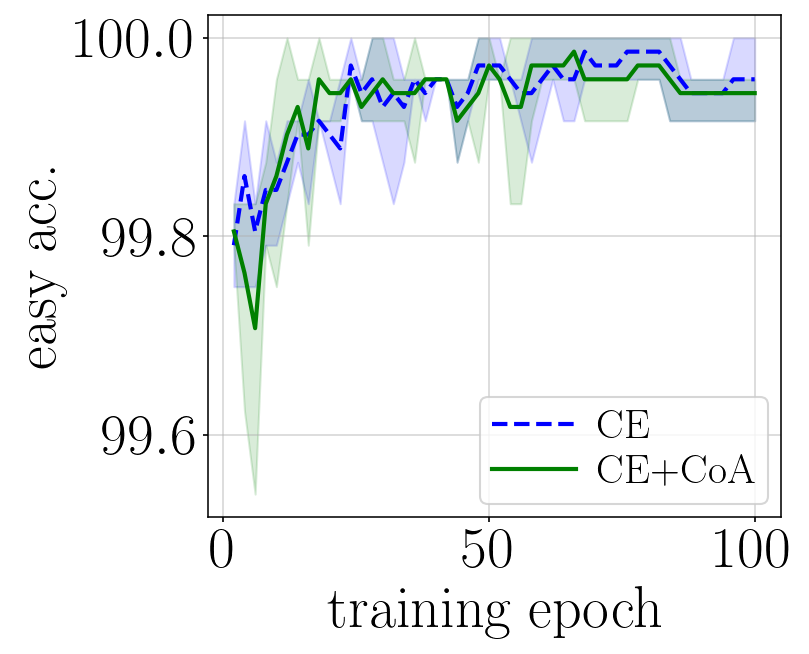}}
    % \subfigure[]{\label{fig:coaloss_confusing}\includegraphics[width=40mm]{resources/CoAloss_confusing.png}}
    \subfigure[]
    {\label{fig:coaloss_confusing}\includegraphics[width=40mm]{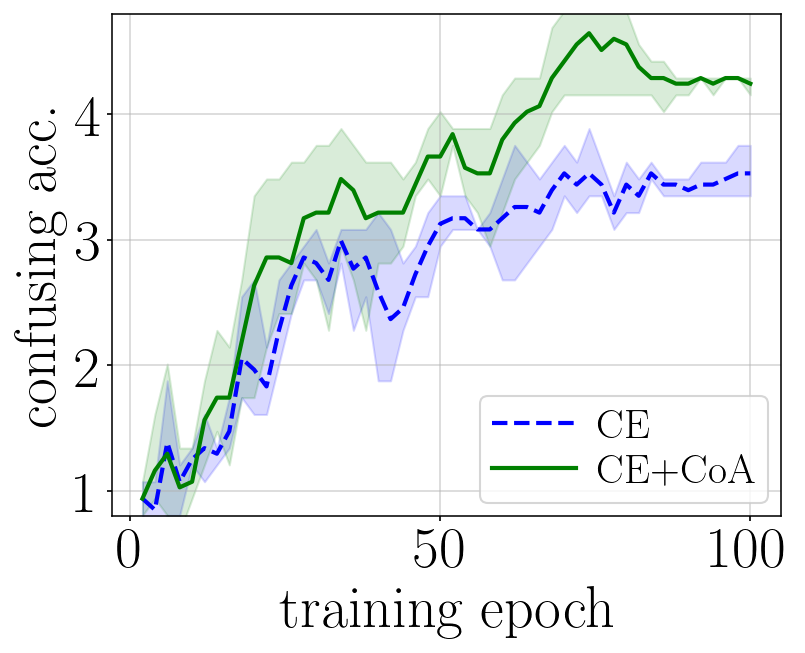}}
    \subfigure[]{\label{fig:coaloss_test}\includegraphics[width=40mm]{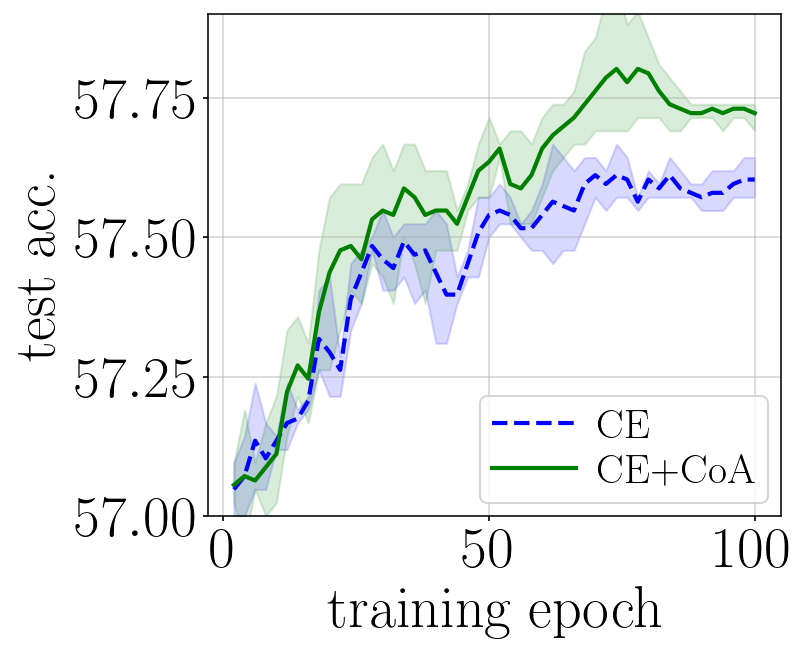}}
    % \caption{(a) Average gradient norm. (b) Accuracy on easy test samples correctly predicted by zero-shot CLIP. (c) Accuracy on confusing test samples misclassified by zero-shot CLIP with a probability gap below $0.5$. (d) Accuracy on all test samples.}\label{fig:coaloss}
    \caption{(a) Proportion of predictions by zero-shot CLIP on EuroSAT. (b) Accuracy on easy test samples correctly predicted by zero-shot CLIP. (c) Accuracy on confusing test samples misclassified by zero-shot CLIP with a probability gap below $0.2$. (d) Accuracy on all test samples.}\label{fig:coaloss}
\end{figure*}

% We evaluate transfer performance across datasets by training on ImageNet with $1,000$ classes in a $4$-shot setting and evaluating on $10$ different datasets with non-overlapping classes.
We evaluate cross-dataset transfer by training on ImageNet with $1,000$ classes in a $4$-shot setting and testing on $10$ different datasets. % with non-overlapping classes.
\cref{tab:transfer} shows the accuracy and harmonic mean for source and target datasets.
% CoOp improves source accuracy over zero-shot CLIP but suffers from reduced target accuracy, highlighting the need for generalization. 
CoOp improves source accuracy over zero-shot CLIP but reduces target accuracy, highlighting the need for generalization. 
ProGrad and KgCoOp similarly enhance source accuracy but fail to outperform zero-shot CLIP on the target dataset, indicating limited transferability of prompt knowledge. 
MaPLe and DePT improve transfer performance by introducing additional parameters but show limited source accuracy. % improvements on the source dataset.
% CoCoA-Mix achieves the highest performance on both source and target datasets, demonstrating that adjusting the confidence of specialized and generalized prompts enhances performance across diverse datasets.
% CoCoA-Mix outperforms existing methods on the source dataset, highlighting the effectiveness of CoA-loss. 
% \TBU{
% On the target dataset, it achieves the second-best performance after MaPLe. 
% The result is competitive because CoCoA-Mix trains only about $0.26\%$ of the parameters compared with MaPLe.
% }
\EDITED{
CoCoA-Mix achieves the highest performance on both source and target datasets, demonstrating effective knowledge transfer across different datasets.
Notably, CoCoA-Mix outperforms existing methods in generalization while utilizing only $0.26\%$ of the parameters used by MaPLe.
}
% Source performance reflects specialization to the training dataset, while target performance represents the average generalization across the 10 datasets.
% The proposed method achieves the highest performance on both source and target datasets, demonstrating its superiority over existing approaches.
% Detailed results on 10 datasets are provided in~\cref{sec:appendix-transfer}.
\cref{sec:appendix-transfer} provides detailed results on 10 datasets.

%-------------------------------------------------------------------------
\subsection{Effectiveness of Our Method}

\begin{table*}[t]
\caption{
Comparison of performance based on the loss function used to train prompts and the strategy of mixing prompts. $\text{T}_\text{En}$ refers to the tuning ensemble methods~\cite{lu2024beyond}. ViT-B/16 shows the average performance over 11 datasets, while RN50+RN101+ViT-B/16+ViT-B/32 reports the average over 10 datasets.
}\label{tab:ensemble_strategy}
\centering
\begin{adjustbox}{width=\textwidth}
\begin{tabular}{p{4cm} ccc ccc}
    \specialrule{1.2pt}{0pt}{1pt}
    Backbone & Method & Loss & Ensemble & \text{Base} & \textit{New} & H \\
    \specialrule{1.2pt}{1pt}{1pt}
    \multicolumn{7}{@{}c@{}}{\rule{0pt}{3pt}}\\[-10pt]
    \multirow{5}{*}{ViT-B/16} 
        & CLIP & - & - & 65.1 & 68.8 & 66.8 \\
        & CLIP (w/ Ensemble) & - & Uniform Ensemble & 70.6 & 74.3 & 72.3 \\
        & CoA-loss (w/o Ensemble) & CoA-loss & - & 78.6 & 68.5 & 72.9 \\
        & CoA-loss (w/ Ensemble) & CoA-loss & Uniform Ensemble & 79.1 & 73.7 & 76.2 \\
        & CoCoA-Mix & CoA-loss & CoA-weights & \textBF{79.3} & \textBF{75.1} & \textBF{77.0} \\[2pt]
    \hline
    \multicolumn{7}{@{}c@{}}{\rule{0pt}{3pt}}\\[-8pt]
    \multirow{3}{=}{RN50 + RN101 \\ + ViT-B/16 + ViT-B/32} 
        & $\text{T}_\text{En}$ + CoCoOp & CoCoOp & Sample-Aware Weight Generator& 84.1 & 75.5 & 79.2 \\
        & $\text{T}_\text{En}$ + CoA-Loss & CoA-loss & Sample-Aware Weight Generator& 85.3 & 75.2 & 79.5 \\
        & CoCoA-Mix & CoA-loss & CoA-weights & \textBF{85.4} & \textBF{76.3} & \textBF{80.3} \\
    \specialrule{1.2pt}{1pt}{0pt}
\end{tabular}
\end{adjustbox}
\end{table*}
%%%%%%%%%%%%%%%%%%%%%%%%%%%%%%%%%%%%%%%%
\paragraph{Performance Comparison with Existing Loss Functions}

We compared {\Base} performance across various loss functions to demonstrate the effectiveness of CoA-loss for specialization in prompt tuning.
\cref{fig:lossfunctions} shows the average {\Base} performance for each loss function.
% Focal loss (FL)~\cite{ross2017focal} emphasizes harder samples but suppresses well-classified ones, limiting optimization in few-shot settings and underperforming compared with CE.
% Generalized cross entropy (GCE)~\cite{zhang2018generalized} generalizes CE and mean absolute error (MAE) to improve robustness in noisy labels but restricts specialization by limiting optimization on misclassified samples. 
Focal loss (FL)~\cite{ross2017focal} emphasizes misclassified samples but struggles to learn from well-classified ones, leading to poor performance compared with CE in prompt tuning.
Generalized cross entropy (GCE)~\cite{zhang2018generalized}, which generalizes CE and mean absolute error (MAE), limits optimization on misclassified samples, restricting specialization. 
CE achieves better performance in prompt tuning by balancing well-classified and misclassified samples.
% MAE treats all samples equally, ensuring consistent learning for predictable ones and slightly enhancing CE performance when combined.
MAE combined with CE treats all samples more equally and ensures consistent learning even in the few-shot setting, slightly improving performance.
However, these loss functions fail to address confusing cases that significantly affect decision boundaries, limiting their performance.
Adding CoA-loss to CE achieves the best performance by effectively focusing on confusing cases.
Detailed results and analysis are provided in~\cref{sec:appendix-lossfunctions}.

%%%%%%%%%%%%%%%%%%%%%%%%%%%%%%%%%%%%%%%%
\paragraph{Effectiveness of Confusion-Aware Loss}

We analyze the effect of CoA-loss on model predictions using the EuroSAT dataset.
\cref{fig:coaloss} compares training progress with and without CoA-loss.
% \cref{fig:coaloss_gradnorm} shows the average gradient norm per epoch, where CE alone causes gradients to diminish to very small values. 
% Given that the prompt is optimized in \texttt{torch.float16}, this scale is sufficiently small to hinder effective optimization.
% In contrast, CoA-loss maintains effective prompt optimization by utilizing even easily predictable samples.
\EDITED{
\cref{fig:coaloss_ratio} shows the proportion of predictions by zero-shot CLIP.  
Easy samples are correctly classified, while confusing samples are misclassified with a probability gap of less than 0.2 between the correct and incorrect classes.  
Hard samples are misclassified with a larger probability gap.
}
% \cref{fig:coaloss_correct} and \cref{fig:coaloss_confusing} present performance on easy samples correctly predicted by zero-shot CLIP and confusing samples misclassified with a probability gap below 0.2.
\cref{fig:coaloss_correct} and \cref{fig:coaloss_confusing} present performance on easy samples and confusing samples.
While CoA-loss performs similarly to CE on easy samples, it significantly improves accuracy on confusing ones.
\cref{fig:coaloss_test} shows that CoA-loss outperforms CE on test data by addressing confusing cases, enhancing class decision boundaries.
The performance improvement on confusing samples with CoA-loss for each dataset is provided in~\cref{sec:appendix-coaloss}

%%%%%%%%%%%%%%%%%%%%%%%%%%%%%%%%%%%%%%%%
\paragraph{Effectiveness of Confidence-Aware Weights}

We evaluate the effectiveness of CoA-weights in the context of prompt ensembling. As shown in~\cref{tab:ensemble_strategy}, we compare various ensembling strategies and loss functions across both single- and multi-backbone configurations. 
In the ViT-B/16 setting, uniform ensembling improves generalization performance over the CLIP baseline by aggregating predictions from multiple prompts. CoCoA-Mix, which integrates CoA-loss and CoA-weights, outperforms all other variants. This highlights the importance of confidence-aware weighting.
In the model-level ensemble setting, CoCoA-Mix achieves the best performance, outperforming the tuning ensemble ($T_\text{En}$) with a sample-aware weight generator~\cite{lu2024beyond}. Notably, CoA-weights achieve this with only two learnable parameters, compared with over 205,204 parameters required by the sample-aware weight generator.

\section*{Conclusion}
\label{sec/5_conclusion}

We proposed CoCoA-Mix, a framework combining CoA-loss and \vtwo{CoA-weights} to enhance specialization and generalization in prompt tuning. 
CoA-loss improves specialization by addressing confusing cases, while \vtwo{CoA-weights} adjust the confidence of predictions to enhance generalization without sacrificing specialization.
We believe our method sets a new direction for effective prompt tuning.

\section*{Acknowledgements}

% \textcolor{red}{\textbf{Do not} include acknowledgements in the initial version of the paper submitted for blind review.}
This work was supported by Korea Evaluation Institute Of Industrial Technology (KEIT) grant funded by the Korea government(MOTIE) (No.20023455, Development of Cooperate Mapping, Environment Recognition and Autonomous Driving Technology for Multi Mobile Robots Operating in Large-scale Indoor Workspace). This research was supported by the KAIST Convergence Research Institute Operation Program.

% If a paper is accepted, the final camera-ready version can (and
% usually should) include acknowledgements.  Such acknowledgements
% should be placed at the end of the section, in an unnumbered section
% that does not count towards the paper page limit. Typically, this will 
% include thanks to reviewers who gave useful comments, to colleagues 
% who contributed to the ideas, and to funding agencies and corporate 
% sponsors that provided financial support.

\section*{Impact Statement}

Prompt tuning has emerged as a promising method for VLMs, enabling efficient adaptation to diverse tasks by optimizing a small set of parameters while keeping the core model frozen. However, real-world scenarios with unseen classes require more generalized solutions. We proposed a mixture model-based approach to enhance generality and improve performance in practical applications. We believe the outcomes of this research are largely positive, making it unnecessary to highlight specific negative impacts in this paper.


\bibliography{main}
\bibliographystyle{icml2025}

%%%%%%%%%%%%%%%%%%%%%%%%%%%%%%%%%%%%%%%%%%%%%%%%%%%%%%%%%%%%%%%%%%%%%%%%%%
% APPENDIX
%%%%%%%%%%%%%%%%%%%%%%%%%%%%%%%%%%%%%%%%%%%%%%%%%%%%%%%%%%%%%%%%%%%%%%%%%%
\newpage
\appendix
\onecolumn

\section{Proof of~\eqref{thm:expected_error_of_mixture_hypothesis}}\label{sec:appendix-proof-main_them}

% \begin{proof} 
Consider $K+1$ individual prompts $\mathcal{T}=\{\bt_0, \bt_1,\dots,\bt_K\}$ and a mixture model $\cocoamix$ with non-negative weights $\bpi=\{\pi_0, \pi_1,\dots,\pi_K\}$, where $\sum_{i=0}^K\pi_i=1$. 
Let $\mathcal{D}_T$ be an arbitrary target domain. 
The expected error $\err{T}{\cocoamix}$ of the mixture model on the target domain is defined as follows in terms of the Kullback-Leibler (KL) divergence:

\newcommand{\E}[1]{\mathbb{E}_{(\bx, y)\sim\mathcal{D}_{#1}}}

\begin{equation*}
    \err{T}{\cocoamix}  = \E{T}\left[ -\log{\cocoamix(y)} \right],
\end{equation*}

\noindent where $y$ is the ground-truth label for the image $\bx$.

Using the definition of the mixture model, $\cocoamix(y) = \softmaxt{\sum_{i=0}^K\pi_i\Sim_{\bt_i}}{\tau}{y}$, the expected error can be decomposed into two terms as follows:

\begin{equation*}\label{eq:proof-main_them1}
\begin{aligned}
    \err{T}{\cocoamix} 
    & = \E{T}\left[ -\log{\cocoamix(y)} \right] &\\
    & = \E{T}\left[ -\log{
            \softmaxt{\sum_{i=0}^K\pi_i\Sim_{\bt_i}}{\tau}{y}
        } \right] 
        &\\
    & = \E{T}\left[ 
            -\sum_{i=0}^K\pi_i\Sim_{\bt_i}(y)/\tau
            +\log{\sum_{l'\in\mathcal{Y}}{\exp\left(
                \sum_{i=0}^K\pi_i\Sim_{\bt_i}(l')/\tau
            \right)}}
        \right] 
        &\\
    & = \begin{aligned}[t]
        & \E{T}\left[
            - \sum_{i=0}^K\pi_i\Sim_{\bt_i}(y)/\tau
            + \sum_{i=0}^K\pi_i{\log{
                \sum_{l'\in\mathcal{Y}}\exp\left(\Sim_{\bt_i}(l')/\tau\right)
            }} \right] \\
        & +\E{T}\left[
            - \sum_{i=0}^K\pi_i{\log{
                \sum_{l'\in\mathcal{Y}}\exp\left(\Sim_{\bt_i}(l')/\tau\right)
            }}
            + \log{\sum_{l'\in\mathcal{Y}}{\exp\left(
                \sum_{i=0}^K\pi_i\Sim_{\bt_i}(l')/\tau
            \right)}}\right].
        \end{aligned}
        &\\
\end{aligned}
\end{equation*}

% The first term is rewritten using the definition of the individual predictive distribution $\pred_{\bt_i}$ for the prompt $\bt_i$, given as $\pred_{\bt_i}(y|\bx)=\softmaxt{\Sim_{\bt_i}}{\tau}{y}$, as follows:
The first term is rewritten using the definition of the individual predictive distribution $\pred_{\bt_i}$ for the prompt $\bt_i$, given as $\pred_{\bt_i}(y)=\softmaxt{\Sim_{\bt_i}}{\tau}{y}$, as follows:

\begin{equation*}\label{eq:proof-main_them2}
\begin{aligned}
    & \E{T}\left[
        - \sum_{i=0}^K\pi_i\Sim_{\bt_i}(y)/\tau
        + \sum_{i=0}^K\pi_i{\log{
            \sum_{l'\in\mathcal{Y}}\exp\left(\Sim_{\bt_i}(l')/\tau\right)}} 
    \right]\\
    & = \E{T}\left[
        -\sum_{i=0}^K\pi_i \left(
                \Sim_{\bt_i}(y)/\tau
                -\log{\sum_{l'\in\mathcal{Y}}\exp\left(\Sim_{\bt_i}(l')/\tau\right)}
            \right)
        \right]\\
    & = \E{T}\left[
        -\sum_{i=0}^K\pi_i \left(
                \log\exp\left(\Sim_{\bt_i}(y)/\tau\right)
                -\log{\sum_{l'\in\mathcal{Y}}\exp\left(\Sim_{\bt_i}(l')/\tau\right)}
            \right)
        \right]\\
    & = \sum_{i=0}^K\pi_i \E{T}\left[
            -\log \softmaxt{\Sim_{\bt_i}}{\tau}{y}
        \right]\\
    & = \sum_{i=0}^K\pi_i \err{T}{\pred_{\bt_i}}.\\
\end{aligned}
\end{equation*}

As a result, the first term is equivalent to a convex combination of the expected errors of the individual predictive distributions with weights $\bpi$.

For the second term, Jensen's inequality~\cite{jensen1906fonctions} can be applied to bound it, as $\log\sum\exp$ is a convex function:

\begin{equation*}\label{eq:proof-main_them3}
\begin{aligned}
    & \E{T}\left[
        - \sum_{i=0}^K\pi_i{\log{
            \sum_{l'\in\mathcal{Y}}\exp\left(\Sim_{\bt_i}(l')/\tau\right)}}
        + \log{\sum_{l'\in\mathcal{Y}}{\exp\left(
            \sum_{i=0}^K\pi_i\Sim_{\bt_i}(l')/\tau\right)}}
    \right] \\
    & \leq \E{T}\left[
        - \sum_{i=0}^K\pi_i{\log{
            \sum_{l'\in\mathcal{Y}}\exp\left(\Sim_{\bt_i}(l')/\tau\right)}}
        + \sum_{i=0}^K\pi_i \left(
            \log{\sum_{l'\in\mathcal{Y}}{\exp\left(
                \Sim_{\bt_i}(l')\tau\right)}}\right)
    \right]\\
    & \leq 0.
\end{aligned}
\end{equation*}

By combining the results from the first and second terms, we conclude that the expected error of the mixture model on the target domain is bounded as follows:

\begin{equation*}
    \err{T}{\cocoamix} \leq \sum_i\pi_i\err{T}{\pred_{\bt_i}}.
\end{equation*}
% \end{proof}

% %%%%%%%%%%%%%%%%%%%%%%%%%%%%%%%%%%%%%%%%%%%%%%%%%%%%%%%%%%%%%%%%%%%%%%%%%%%%%%%
% %%%%%%%%%%%%%%%%%%%%%%%%%%%%%%%%%%%%%%%%%%%%%%%%%%%%%%%%%%%%%%%%%%%%%%%%%%%%%%%

% ------------------------------------------------------------ %
\section{Statistical Validation of~\cref{ass:divergence_class}}\label{sec:appendix-ass_experiments}

To empirically verify~\cref{ass:divergence_class}, we conducted a statistical experiment using the CIFAR-100~\cite{krizhevsky2009learning} dataset and the CLIP model. Specifically, we randomly partitioned the $100$ classes into $50$ in-classes and $50$ out-classes. We then trained specialized prompts $\bt_i$ on the in-class subset using prompt tuning and compared their predictions against the zero-shot CLIP baseline on both domains. This process was repeated over $10$ random splits.

\begin{figure}[H]
    \centering
    \includegraphics[width=0.90\textwidth]{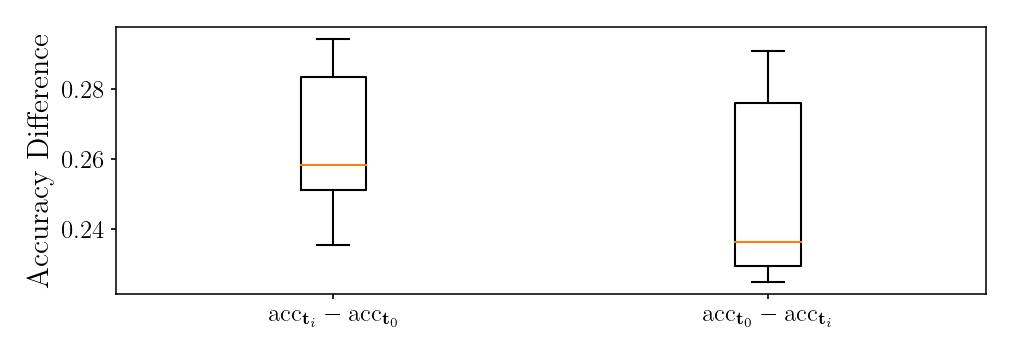}
    \caption{
    Box plots of accuracy differences across $10$ random splits. The left shows performance gains of specialized over generalized predictions on in-class domain $\mathcal{D}_{T_i}$, i.e. $\text{acc}_{\bt_i} - \text{acc}_{\bt_0}$. The right shows gains of the generalized prediction on out-class domain $\mathcal{D}_{T_{j\neq i}}$, i.e. $\text{acc}_{\bt_0} - \text{acc}_{\bt_i}$.
    }
    \label{fig:statistic_experiment}
\end{figure}

\cref{fig:statistic_experiment} presents a box plot summarizing the performance differences between the specialized prompt $\bt_i$ on the in-class domain $\mathcal{D}_{T_i}$ and the generalized zero-shot prompt $\bt_0$. The results show that $\bt_i$ consistently outperforms $\bt_0$ on in-class samples from $\mathcal{D}_{T_i}$, whereas $\bt_0$ achieves higher accuracy on out-class samples from $\mathcal{D}_{T_{j\neq i}}$.

To assess statistical significance, we performed one-sided paired $t$-tests on the per-split accuracy gaps. The resulting $p$-values were $9.25\times 10^{-12}$ for the in-class domain comparison ($\text{acc}_{\bt_i} > \text{acc}_{\bt_0}$) and $2.06\times 10^{-10}$ for the out-class domain comparison ($\text{acc}_{\bt_0}>\text{acc}_{\bt_i}$), both significantly below the standard threshold of $0.05$. These results allow us to reject the null hypothesis and confirm that both inequalities in~\cref{ass:divergence_class} hold with strong statistical confidence.

These findings support the assumption that specialized predictions are more effective within their domain, whereas generalized predictions are preferable for unseen class domains.

% ------------------------------------------------------------ %
\newpage
\section{Effect of Cross-Entropy in the Mixture Model}\label{sec:appendix-effect_ce}

The derivative of the cross-entropy $\mathcal{L}_\text{CE}$ for the mixture model $\cocoamix$ with respect to $\pi_i^\text{in}$ is as follows:

\newcommand{\piin}{{\pi_i^\text{in}}}

\begin{equation*}\begin{aligned}
    \frac{\partial\mathcal{L}_\text{CE}(\bx,y;\cocoamix)}{\partial{\pi_i}}
    & = \frac{\partial \left( -\log\cocoamix(y) \right)}{\partial{\pi_i}}\\
    & = \frac{-1}{\cocoamix(y)}\frac{\partial\cocoamix(y)}{\partial{\pi_i}}\\
    & = \frac{-1}{\cocoamix(y)}\frac{\partial}{\partial{\pi_i}}\Bigg(
            \frac{\exp\left( 
                \sum_{i=0}^K \pi_i \Sim_{\bt_i}(y)/\tau
            \right) }
            {\sum_{l'\in\mathcal{Y}} \exp\left( 
                \sum_{i=0}^K \pi_i \Sim_{\bt_i}(l')/\tau
            \right) }
        \Bigg)\\
    & = \frac{-1}{\cocoamix(y)}\left(
            \Sim_{\bt_i}(y)\cocoamix(y)
            - \cocoamix(y)\sum_{l\in\mathcal{Y}}\cocoamix(l)\Sim_{\bt_i}(l)
        \right)/\tau\\
    & = -\left( 
            \Sim_{\bt_i}(y) - \sum_{l\in\mathcal{Y}}\cocoamix(l)\Sim_{\bt_i}(l)
        \right)/\tau\\
    & = -\left( 
            \Sim_{\bt_i}(y) - \mathring{\Sim}_{\bt_i}
        \right)/\tau,
\end{aligned}\end{equation*}

where $\mathring{\Sim}_{\bt_i}$ is the importance-weighted similarity defined as a weighted sum of the predicted probability of the mixture model and the similarity derived from the prompt $\bt_i$, 
i.e. $\mathring{\Sim}_{\bt_i}=\sum_{l\in\mathcal{Y}}{\cocoamix(l)\Sim_{\bt_i}(l)}$.
For example, if the mixture model predicts class $l^*$ with the highest probability, $\mathring{\Sim}_{\bt_i}$ approximates the similarity $\Sim_{\bt_i}(l^*)$ for class $l^*$ derived from prompt $\bt_i$. 
Here, we explain how the CoA-weights $\pi_i$ for in-classes is optimized through the cross-entropy of the mixture model.
For simplicity, we assume $\mathring{\Sim}_{\bt_i}\approx \Sim_{\bt_i}(l^*)$, where $l^*=\argmax_l\cocoamix(l)$.

In the case $\Sim_{\bt_i}(y) > \mathring{\Sim}_{\bt_i}$, the prompt $\bt_i$ predicts the correct class $y$ with high similarity.
% Therefore, when the mixture model misclassifies, i.e., $l^*\neq y$, the hand-crafted prompt $\bt_0$ provides low similarity for the correct class $y$. 
Therefore, when the mixture model misclassifies, i.e., $l^*\neq y$, the other prompts $\bt_{j\neq i}$ provide low similarities for the correct class $y$. 
This case results in an increase in $\pi_i$ through gradient updates, encouraging the mixture model to rely more on $\bt_i$.

Conversely, if $\Sim_{\bt_i}(y) < \mathring{\Sim}_{\bt_i}$, the prompt $\bt_i$ predicts the correct class $y$ with low similarity. 
When the mixture model correctly classifies, i.e. $l^*=y$, it suggests that the other prompts $\bt_{j\neq i}$ provide high similarities for the correct class $y$, while the prompt $\bt_i$ underperforms. This case decreases $\pi_i$, allowing the mixture model to trust the other prompts $\bt_{j\neq i}$ more.

\newpage

% ------------------------------------------------------------ %
\section{Implementation Details}\label{sec:appendix-implementation_details}

% ==== % 
\subsection{Details of CoA-Weights Optimization} \label{sec:implementation_details-coaweights}

There exist multiple strategies for optimizing the CoA-weights $\bpi$. 
To evaluate the effectiveness of different optimization strategies for CoA-weights, we compared two implementation approaches: one-stage optimization and two-stage optimization. 
Specifically, for one-stage optimization, we design the optimization process such that the temperature scale is weighted by $\pi_i$, i.e., $\tau_i = \pi_i^{-1} \tau$. We fix $\tau_0$ to the temperature of the pre-trained CLIP model, i.e. $\tau_0=0.01$, and optimize $\tau_1$ jointly with prompt parameters. 
This enables the computation of $\bpi$ satisfying $\sum_{i=0}^K \pi_i = 1$, ensuring compatibility with the standard temperature scaling framework of CLIP. The optimization of a single scalar $\tau_1$ suffices to determine both $\bpi$ and the global temperature $\tau$, via the relation $\tau_i = \pi_i^{-1} \tau$. 
However, the temperature scale $\tau$ is not fixed during optimization, which may introduce instability when $K\geq 2$.
In contrast, the two-stage strategy decouples prompt tuning and CoA-weights optimization. 
Here, $\bpi$ is parameterized using a softmax of parameters $\alpha_0,\cdots,\alpha_K$. Concretely, we fix the pre-softmax logit $\alpha_0 = 0$ and optimize $\alpha_i$ under a fixed temperature $\tau = 0.01$. 
The comparison of these strategies for the $K=1$ case is reported in~\cref{tab:optimization}. The two strategies yield comparable performance, suggesting that the proposed loss function is robust to the specific choice of optimization parameterization for CoA-weights.
In this paper, we used the one-stage strategy for most benchmarks and adopted the two-stage approach specifically for FSCIL tasks due to its stability advantages.

\begin{table}[b!]
\caption{
Comparison of various optimization strategies with respect to $\pi_1$ and $\tau$ for \textit{Base} and \textit{New} domains. In one-stage optimization, $\tau_0$ is set to the temperature scale from pre-trained CLIP, i.e. 0.01. $\bpi=\{\pi_0,\pi_1\}$ and $\tau$ are determined by optimizing $\tau_1$. In two-stage optimization, $\alpha_0$ is set to $0$, and $\bpi$ and $\tau$ are determined by optimizing $\alpha_1$. 
}\label{tab:optimization}
\centering
\begin{tabular}{lcccccc}
\toprule
% & $\alpha$ & $\tau$ & \textit{Base} & \textit{New} & H \\
 & $\pi_1$ & $\tau$ & \textit{Base} & \textit{New} & H \\
\midrule
One-Stage Optimization & 
$\frac{\tau_0}{\tau_1 + \tau_0}$ & 
$\frac{\tau_1 \cdot \tau_0}{\tau_1 + \tau_0}$ & 
79.3 & \textbf{75.1} & \textbf{77.0} \\
Two-Stage Optimization & 
$\frac{\exp(\alpha_1)}{\exp(\alpha_0) + \exp(\alpha_1)}$ & 
$0.01$ & 
\textbf{79.4} & 75.0 & \textbf{77.0} \\
\bottomrule
\end{tabular}
\end{table}

% ==== % 
\subsection{Details of Dataset} \label{sec:details_of_datasets}

\begin{table*}[b!]
    \caption{
    11 datasets used for base-to-new generalization and cross-dataset transfer
    } \label{tab:appendix-datasets}
    \begin{center}
        \begin{threeparttable}
        \renewcommand{\arraystretch}{1.1}
        \footnotesize{
            \begin{tabular}
                % {@{}r | *{1}{c}@{} | *{3}{c}@{}}
                {@{}r | m{1.0cm} m{2.5cm} m{5.0cm} m{4.0cm}@{}}
                \toprule
                 \multicolumn{1}{c|@{}}{Dataset}
                 & \mc{\MethodFont{Classes}} 
                 & \mc{\MethodFont{Task}}
                 & \mc{\MethodFont{Description}}
                 & \mc{\MethodFont{Example Classes}} 
                \\
                \midrule
                % ------------------------------------ %
                \mcl{\MethodFont{ImageNet}}
                & \mc{1,000} 
                & Object recognition 
                & Large-scale dataset for object classification with diverse categories
                & tench, goldfish, great white shark, a tiger shark, etc. \\
                % ------------------------------------ %
                \mcl{\MethodFont{Caltech101}} 
                & \mc{100} 
                & Object recognition
                & Variety of object categories with random background images.
                & Accordion, Airplane, Brain, Butterfly, Crab, Motorbike, etc. \\
                % ------------------------------------ %
                \mcl{\MethodFont{OxfordPets}} 
                & \mc{37} 
                & Fine-grained object recognition
                & Classification of pet breeds including cats and dogs.
                & Bengal, Persian, Beagle, American Bulldog, etc. \\
                % ------------------------------------ %
                \mcl{\MethodFont{StanfordCars}} 
                & \mc{196} 
                & Fine-grained object recognition 
                & Images of various vehicle types, annotated by model.
                & 2000 AM General Hummer SUV, 2007 BMW X5 SUV, etc.\\
                % ------------------------------------ %
                \mcl{\MethodFont{Flowers102}} 
                & \mc{102} 
                & Fine-grained object recognition 
                & Classification of various flower species.
                & Daffodil, Pink Primrose, Tiger Lily, Yellow Iris, etc. \\
                % ------------------------------------ %
                \mcl{\MethodFont{Food101}} 
                & \mc{101} 
                & Object recognition 
                & User-uploaded real-world food photos with varied backgrounds and noise.
                & Apple Pie, Waffles, Sushi, Chocolate Cake, Bibimbap, etc. \\
                % ------------------------------------ %
                \mcl{\MethodFont{FGVCAircraft}} 
                & \mc{100} 
                & Fine-grained object recognition 
                & Aircraft images annotated hierarchically by variant, family, and manufacturer.
                & Boeing 717, DH-82, Falcon 2000, etc. \\
                % ------------------------------------ %
                \mcl{\MethodFont{SUN397}} 
                & \mc{397}
                & Scene recognition 
                & Covers diverse scenes including indoor, urban, and natural environments.
                & Abbey, Airport Terminal, Bedroom, Harbor, Bar, etc. \\
                % ------------------------------------ %
                \mcl{\MethodFont{DTD}} 
                & \mc{47} 
                & Texture attribute recognition 
                & Real-world texture images annotated with descriptive attributes.
                & Striped, Dotted, Cracked, Fibrous, Scaly, Zigzagged, etc. \\
                % ------------------------------------ %
                \mcl{\MethodFont{EuroSAT}} 
                & \mc{10} 
                & Land use and land cover classification
                & Satellite images from Sentinel-2 focusing on land use and cover types.
                & Annual Crop Land, Forest, Highway or Road, River, etc.\\
                % ------------------------------------ %
                \mcl{\MethodFont{UCF101}} 
                & \mc{101} 
                & Action recognition 
                & Video clips of human actions collected from YouTube in dynamic, real-world environments.
                & Apply Eye Makeup, Basketball Dunk, Playing Piano, etc. \\
                \bottomrule
            \end{tabular}
        }
        \end{threeparttable}
    \end{center}
\end{table*}

\cref{tab:appendix-datasets} lists datasets for base-to-new generalization and cross-dataset transfer. Evaluation across 11 datasets highlights generalizability and efficiency beyond specific tasks or domains.

% ==== % 
\subsection{Base-to-New Generalization} \label{sec:implementation_details-base2new}

Experiments were performed utilizing CLIP with a ViT-B/16~\cite{dosovitskiy2020image} backbone. 
Training was conducted over $50$ epochs with a batch size of $32$. The prompt $\bt$ was optimized using the Adam optimizer with a learning rate of $0.002$ and a weight decay of $5\times 10^{-4}$. CoA-weights were optimized using SGD with the same learning rate, a momentum of $0.9$, and a weight decay of $5\times 10^{-4}$. \vcase{
The weight for $\mathcal{L}_\text{CoA}$ was set to $w=5.0$, the weight for $\mathcal{L}_\text{Ent}$ was set to $10.0$, and the margin was set to $d=0.2$. The prompt length $M$ was set to $16$.
}{
The weight for $\mathcal{L}_\text{CoA}$ was set to $w=5.0$, the weight for $\mathcal{L}_\text{Ent}$ was set to $8.0$, and the margin was set to $d=0.2$. The prompt length $M$ was set to $16$.
}

% ==== % 
\subsection{Few-Shot Class-Incremental Learning} \label{sec:implementation_details-fscil}

Following Ran et al.~\yrcite{ran2024brain}, experiments were conducted using CLIP with a ViT-L/14~\cite{dosovitskiy2020image} backbone.
CoCoA-Mix used prompts of length $M=2$ per session and accumulated them across sessions, requiring fewer parameters than the baseline except in the final two sessions.
Each prompt was trained for specialization within its session, and the final prediction used all prompts from previous sessions.
%%% v1
% CoA-weights were re-trained in each session and applied to all prompts.
Tuned prompt $\bt_i$ for each session, along with CoA-weights $\pi_i^\text{in}$ and $\pi_i^\text{out}$, were stored and reused. To ensure scaling stability across sessions, we used the two-stage optimization strategy for CoA-weights. Considering the number of iterations per session, CoA-weights were optimized for 2 epochs in the initial session, and for 100 epochs in all subsequent sessions. The margin $d$ of the loss $\mathcal{L}_\text{Ent}$ was set to $0.1$.
In the initial session, the out-class set was generated using random words, and in subsequent sessions, it consisted of classes from previous sessions.
All other settings followed those of base-to-new generalization.
% The initial session was trained for $10$ epochs, while each subsequent session was trained for five epochs.

% ==== % 
\subsection{Cross-Dataset Transfer}\label{sec:implementation_details-transfer}

Experiments used CLIP with a ViT-B/16 backbone.
The weight for $\mathcal{L}_\text{CoA}$ was set to $w=7.0$, while the loss weight for $\pi^\text{in}$ was set to 2.0.
Other settings followed those of base-to-new generalization.

% ------------------------------------------------------------ %
% ------------------------------------------------------------ %
\newpage
% ------------------------------------------------------------ %
% ------------------------------------------------------------ %
\section{Additional Experimental Results}\label{sec:appendix-additional_experimental_results}

\subsection{Cross-Dataset Transfer}\label{sec:appendix-transfer}

\cref{tab:appendix-transfer} presents the accuracy of the method trained on a single source dataset and evaluated on both source and target datasets.  
Parenthesized values represent the performance difference relative to zero-shot CLIP.
CoCoA-Mix achieves the highest accuracy on the source dataset, highlighting the effectiveness of CoA-loss in specialization.  
It also achieved the highest average accuracy across 10 target datasets, validating the effect of CoA-weights.
For FGVCAircraft, DTD, and EuroSAT, where zero-shot CLIP shows significant performance gaps between source and target datasets, CoCoA-Mix exhibits the most stable and effective transferability, maintaining high performance with minimal accuracy degradation compared with previous methods.
This suggests that CoCoA-Mix is particularly effective under significant domain shifts.
The relatively lower generalization improvement over base-to-new generalization is discussed in~\cref{sec:appendix-limitations}.

\newcommand{\up}[1]{\textcolor{blue}{\ (#1)}}
\newcommand{\down}[1]{\textcolor{red}{\ (#1)}}
\begin{table*}[t!]
    \caption{Performance comparison on 11 datasets in cross-dataset transfer. 
    } \label{tab:appendix-transfer}
    \begin{center}
        \begin{threeparttable}
        \footnotesize{
            \begin{tabular}
                {@{}r | *{1}{c}@{} | *{5}{c}@{}}
                \toprule\toprule
                 \multicolumn{1}{c|@{}}{\multirow{2}{*}{Method}}
                 & \mcl{\MethodFont{Source}} 
                 & \mcl{\MethodFont{Target}} 
                 & \multicolumn{4}{c@{}}{\MethodFont{Target}}\\
                 & \Datasetonel{ImageNet} 
                 & \mcl{\MethodFont{Average}}
                 & \Datasetone{Caltech101} 
                 & \Datasetone{OxfordPets}
                 & \Datasetone{StanfordCars}
                 & \Datasetone{Flowers102}
                \\
                \midrule
                % ------------------------------------------------------------- %
                \mcl{\MethodFont{CLIP}} & \mcl{66.73} % ImageNet
                % Average       % Caltech101    % OxfordPets    
                & \mcl{64.89}   & \mc{93.27}    & \mc{89.18}
                % StanfordCars  % OxfordFlowers
                & \mc{65.56}    & \mc{68.05} \\
                % ------------------------------------------------------------- %
                \mcl{\MethodFont{CoOp}} 
                & \mcl{69.06} % ImageNet
                & \mcl{59.88 \down{-5.01}} % Average
                & \mc{91.06 \down{-2.21}} % Caltech101
                & \mc{86.74 \down{-2.44}} % OxfordPets
                & \mc{59.84 \down{-5.72}} % StanfordCars
                & \mc{62.38 \down{-5.67}} % OxfordFlowers
                \\
                % ------------------------------------------------------------- %
                \mcl{\MethodFont{ProGrad}} 
                & \mcl{70.21} % ImageNet
                & \mcl{62.36 \down{-2.53}} % Average
                & \mc{92.41 \down{-0.86}} % Caltech101
                & \mc{87.90 \down{-1.28}} % OxfordPets
                & \mc{62.94 \down{-2.62}} % StanfordCars
                & \mc{66.98 \down{-1.07}} % OxfordFlowers
                \\
                % ------------------------------------------------------------- %
                \mcl{\MethodFont{KgCoOp}} 
                & \mcl{70.52} % ImageNet
                & \mcl{64.45 \down{-0.43}} % Average
                & \mc{93.55 \up{+0.28}} % Caltech101  
                & \bmc{89.86 \up{+0.68}} % OxfordPets
                & \mc{65.61 \up{+0.05}} % StanfordCars
                & \mc{68.33 \up{+0.28}} % OxfordFlowers
                \\
                % ------------------------------------------------------------- %
                \mcl{\MethodFont{MaPLe}} 
                & \mcl{69.53} % ImageNet
                & \mcl{65.24 \up{+0.35}} % Average
                & \mc{93.43 \up{+0.16}} % Caltech101  
                & \mc{89.77 \up{+0.59}} % OxfordPets
                & \mc{65.70 \up{+0.14}} % StanfordCars
                & \bmc{71.17 \up{+3.12}} % OxfordFlowers
                \\
                % ------------------------------------------------------------- %
                \mcl{\MethodFont{DePT}} 
                & \mcl{68.03} % ImageNet
                & \mcl{65.06 \up{+0.17}} % Average
                & \bmc{94.07 \up{+0.80}} % Caltech101  
                & \mc{89.43 \up{+0.25}} % OxfordPets
                & \bmc{65.87 \up{+0.31}} % StanfordCars  
                & \mc{69.93 \up{+1.88}} % OxfordFlowers
                \\
                \rowcolor{lightgray!50}
                \mcl{\MethodFont{CoCoA-Mix}} % vit_b16_ep50_v_w7-tin_w2-tout_w10_d02_word
                & \bmcl{\vcase{70.85}{?}} % ImageNet
                & \bmcl{\vcase{65.27 \up{+0.38}}{?}} % Average
                & \mc{\vcase{93.46 \up{+0.19}}{?}} % Caltech101
                & \mc{\vcase{89.07 \down{-0.11}}{?}} % OxfordPets
                & \mc{\vcase{65.59 \up{+0.03}}{?}} % StanfordCars
                & \mc{\vcase{68.72 \up{+0.67}}{?}} % OxfordFlowers
                \\
                \toprule\toprule
                 \multicolumn{1}{c|@{}}{\multirow{2}{*}{Method}}
                 & \multicolumn{6}{c@{}}{Target} \\
                 & \Datasetone{Food101}
                 & \Datasetone{FGVCAircraft}
                 & \Datasetone{SUN397}
                 & \Datasetone{DTD}
                 & \Datasetone{EuroSAT}
                 & \Datasetone{UCF101} \\
                \midrule
                % ------------------------------------------------------------- %
                \mcl{\MethodFont{CLIP}} 
                & \mc{85.43} % Food101
                & \bmc{24.81} % fgvc_aircraft
                & \mc{62.61} % sun397
                & \mc{44.09} % dtd
                & \bmc{48.36} % eurosat
                & \mc{67.51} % ucf101
                \\
                % ------------------------------------------------------------- %
                \mcl{\MethodFont{CoOp}} 
                & \mc{83.29 \down{-2.14}} % Food101
                & \mc{16.71 \down{-8.10}} % fgvc_aircraft
                & \mc{59.40 \down{-3.21}} % sun397
                & \mc{38.44 \down{-5.65}} % dtd
                & \mc{39.24 \down{-9.12}} % eurosat
                & \mc{61.66 \down{-5.85}} % ucf101
                \\
                % ------------------------------------------------------------- %
                \mcl{\MethodFont{ProGrad}} 
                & \mc{84.37 \down{-1.06}} % Food101
                & \mc{17.10 \down{-7.71}} % fgvc_aircraft
                & \mc{62.67 \up{+0.06}} % sun397
                & \mc{39.87 \down{-4.22}} % dtd
                & \mc{45.39 \down{-2.97}} % eurosat
                & \mc{63.98 \down{-3.53}} % ucf101
                \\
                % ------------------------------------------------------------- %
                \mcl{\MethodFont{KgCoOp}} 
                & \mc{85.83 \up{+0.40}} % Food101
                & \mc{21.18 \down{-3.63}} % fgvc_aircraft
                & \mc{64.84 \up{+2.23}} % sun397
                & \mc{44.30 \up{+0.21}} % dtd
                & \mc{44.64 \down{-3.72}} % eurosat
                & \mc{66.39 \down{-1.12}} % ucf101
                \\
                % ------------------------------------------------------------- %
                \mcl{\MethodFont{MaPLe}} 
                & \mc{86.13 \up{+0.70}} % Food101
                & \mc{23.27 \down{-1.54}} % fgvc_aircraft
                & \bmc{66.43 \up{+3.82}} % sun397
                & \mc{44.83 \up{+0.74}} % dtd
                & \mc{43.73 \down{-4.63}} % eurosat
                & \bmc{67.93 \up{+0.42}} % ucf101
                \\
                % ------------------------------------------------------------- %
                \mcl{\MethodFont{DePT}} 
                & \bmc{86.27 \up{+0.84}} % Food101
                & \mc{22.10 \down{-2.71}} % fgvc_aircraft
                & \mc{65.77 \up{+3.16}} % sun397
                & \mc{45.53 \up{+1.44}} % dtd
                & \mc{44.00 \down{-4.36}} % eurosat
                & \mc{67.60 \up{+0.09}} % ucf101
                \\
                \rowcolor{lightgray!50}
                \mcl{\MethodFont{CoCoA-Mix}}  % vit_b16_ep50_v_w7-tin_w2-tout_w10_d02_word
                & \mc{\vcase{85.78 \up{+0.35}}{?}} % Food101
                & \mc{\vcase{24.10 \down{-0.71}}{?}} % fgvc_aircraft
                & \mc{\vcase{63.61 \up{+1.00}}{?}} % sun397
                & \bmc{\vcase{46.41 \up{+2.32}}{?}} % dtd
                & \mc{\vcase{48.18 \down{-0.18}}{?}} % eurosat
                & \mc{\vcase{67.78 \up{+0.27}}{?}} % ucf101
                \\
                \bottomrule\bottomrule
            \end{tabular}
        }
        \end{threeparttable}
    \end{center}
\end{table*}

% ------------------------------------- %
\newpage

\subsection{Performance Improvement on Confusing Samples of CoA-loss}\label{sec:appendix-coaloss}

\begin{figure}[H]
    \centering
    \includegraphics[width=0.50\textwidth]{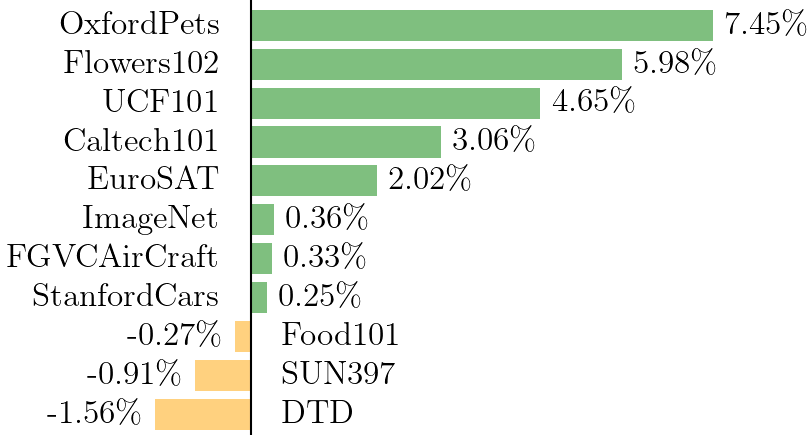}
    \caption{
    Performance improvement on confusing samples, defined as instances with a probability gap $\leq 0.5$ between correct and incorrect labels or misclassified by zero-shot CLIP.
    }
    \label{fig:effect_coa_datasets}
\end{figure}

We define confusing samples as those misclassified by zero-shot CLIP with a probability gap of $0.5$ or less between the correct and incorrect classes. 
\cref{fig:effect_coa_datasets} presents the performance of confusing samples from the test data across 11 datasets for the {\Base} classes.
While slight performance drops are observed on datasets such as Food101, SUN397, and DTD, CoA-loss consistently enhances specialization across most datasets, with notable improvements on OxfordPets, Flowers102, and UCF101.
This demonstrates the effectiveness of CoA-loss in handling confusing cases while enhancing the performance across diverse datasets.

\newpage
% ------------------------------------- %
\subsection{{\Base} Performance Comparison According to Loss Function}\label{sec:appendix-lossfunctions}

\begin{figure*}[b]
    \centering
    \includegraphics[width=0.90\textwidth]{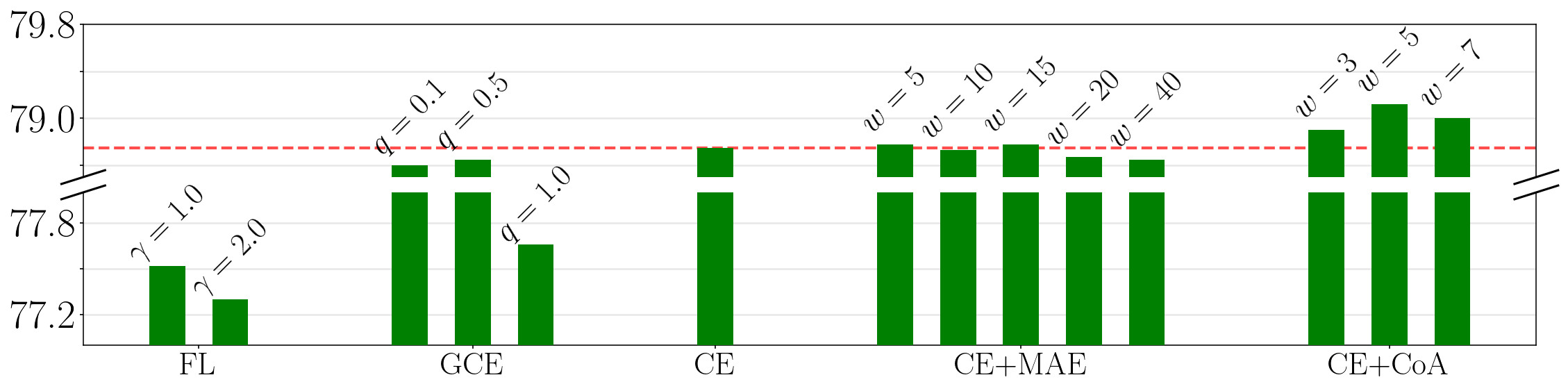}
    \caption{
    Performance comparison across various loss functions with varying hyperparameters. The hyperparameter setting for each loss is shown above each data.
    }
    \label{fig:details_loss_functions}
\end{figure*}

We compared CoA-loss with four loss functions: focal loss (FL)~\cite{ross2017focal}, generalized cross-entropy loss (GCE)~\cite{zhang2018generalized}, cross-entropy (CE), and mean absolute error (MAE).
The results are presented in \cref{fig:details_loss_functions}.

FL modifies CE by weighting with $(1-\pred)^\gamma$ as follows:
\begin{equation}
    \mathcal{L}_\text{FL}(\bx,y;\pred) = - \left(1-\pred(y)\right)^\gamma\log\pred(y),
\end{equation} 
\noindent where $\gamma \geq 0$ is a focusing parameter.
FL prioritizes misclassified samples but limits learning from well-classified ones, leading to lower performance in prompt tuning.

GCE generalizes CE and MAE as follows:
\begin{equation}
    \mathcal{L}_\text{GCE}(\bx,y;\pred) = \frac{1-\pred(y)^q}{q},
\end{equation}
\noindent where $q\in(0,1]$ transitions from CE ($q\rightarrow 0$) to MAE ($q=1$). 
While designed to mitigate the overfitting of CE and slow convergence of MAE, GCE underperforms CE in prompt tuning.

CE, widely used in prompt tuning, is defined as follows:
\begin{equation}
    \mathcal{L}_\text{CE}(\bx,y;\pred) = - \log\pred(y).
\end{equation}
Its gradients with respect to the similarity $\Sim(y)$ and $\Sim(c\neq y)$ are as follows, respectively:
\begin{equation}
    \begin{aligned}
        \frac{\partial \mathcal{L}_\text{CE}}{\partial \Sim(y)} = -\frac{1}{\tau}\left(1-\pred(y)\right) & & \text{ and } & & 
        \frac{\partial \mathcal{L}_\text{CE}}{\partial \Sim(c\neq y)} = \frac{1}{\tau}\pred(c), \\
    \end{aligned}
\end{equation}
\noindent where $\tau$ is the temperature.
CE updates gradients based only on individual class probabilities, ignoring inter-class relationships.

Adding MAE to CE, the loss is defined as follows:
\DeclareRobustCommand{\1}[1]{\text{\usefont{U}{bbold}{m}{n}1}_{#1}}
\begin{equation}
    \mathcal{L}_\text{CE+MAE}(\bx,y;\pred) = -\log\pred(y) + w \frac{1}{|\mathcal{Y}|}\sum_{l\in\mathcal{Y}} \left| \1{l=y}-\pred(l)\right|,
\end{equation}
\noindent where $w$ controls the contribution of MAE. 
While CE emphasizes difficult samples, MAE ensures uniform learning over all samples, improving robustness to noise. 
With optimized $w$, CE+MAE slightly improves performance, but excessive MAE weighting reduces focus on difficult samples, degrading performance.

CoA-loss, similar to MAE, applies only to the correct class $y$ and is combined with CE as follows:
\begin{equation}
    \mathcal{L}_\text{CE+CoA}(\bx,y;\pred) = -\log\pred(y) + w\left(1-\pred(y)\right),
\end{equation}
\noindent where $w$ controls the contribution of CoA-loss. 
Unlike CE, CoA-loss considers both $\pred(c\neq y)$ and $\pred(y)$ when updating $\pred(c\neq y)$, focusing on confusing cases where $\pred(y)$ and $\pred(c\neq y)$ approach 0.5. By focusing learning near decision boundaries, CE+CoA achieves the highest specialization performance.

% \newpage
% % ------------------------------------- %
% \subsection{Performance comparison based on the number of parameters}\label{sec:appendix-parameter_comparison}

% 우리는 \cref{fig:params_performance}에 각 방법론들의 parameters 개수에 따른 specialization과 generalization 성능을 비교하였다.
% 원의 크기는 learnable parameters의 개수에 비례하며 x축은 {\Base}에서의 성능을, y축은 {\New}에서의 평균 성능을 보여준다.
% 별모양은 zero-shot CLIP의 성능을 보여준다.

% CoOp, ProGrad, KgCoOp의 파라미터 개수는 $M\times D$이다.
% MaPLe의 파라미터 개수는 아래와 같이 계산된다:
% \begin{equation}
%     \underbrace{(M\times D)}_\text{learnable prompt}
%     + \underbrace{(D\times D_\text{out} + D_\text{out})}_\text{projection linear layer}
%     + \underbrace{(M\times D\times (J-1))}_\text{compound prompts text}
%     + \underbrace{(D\times D_\text{out} + D_\text{out})\times (J-1)}_\text{compound prompt projection},
% \end{equation}
% where prompt length $M=2$, compound prompts depth $J=9$, embedding dimension $D=512$, $D_\text{out}=768$

% \begin{figure*}[t]
%     \centering
%     \includegraphics[width=0.70\textwidth]{resources/parameter_comparison.png}
%     \caption{
%     Comparison of specialization and generalization performance based on the number of learnable parameters.
%     Larger circles represent more parameters, indicating better performance with smaller circles positioned in the upper right corner.
%     }
%     \label{fig:params_performance}
% \end{figure*}

% \TBU{(TBU)}

% % ------------------------------------- %
% \subsection{Hand-crafted prompt에 따른 성능 민감도}

% \TBU{(TBU)}

\newpage
%%%%%%%%%%%%%%%%%%%%%%%%%%%%%%%%%%%%%%%%
\section{Ablation Studies}\label{sec:appendix-ablation_studies}

%%%%%%%%%%%%%%%%%%%%%%%%%%%%%%%%%%%%%%%%
\subsection{Effectiveness of Confidence-Aware Weights}

\newcommand{\cmark}{\ding{51}}%
\newcommand{\xmark}{\ding{55}}%

% \begin{table}[tb]
% \caption{
% % Impact on optimizing CoA-temp parameters for \textit{Base} ($\tau_v^\text{in}$) and \textit{New} ($\tau_v^\text{out}$) classes.
% Effect of \vtwo{CoA-weight} on {\Base} and {\New} classes.
% } \label{tab:ablation_temp}
% \begin{center}
%     \begin{tabular}
%         {c c c c c}
%         \toprule 
%             % \MethodFont{$\tau_\bv^\text{in}$} 
%             % & \MethodFont{$\tau_\bv^\text{out}$}
%             \vtwo{$\pi_i^\text{in}$} 
%             & \vtwo{$\pi_i^\text{out}$}
%             & \MethodFont{Base}    & \MethodFont{New}    & \MethodFont{H} \\
%         \midrule
%             \xmark & \xmark 
%                 % & 79.25 & 73.90 & 76.32\\
%                 & 79.12 & 73.66 & 76.15\\ % CoCoA-Mix
%             \cmark & \xmark
%                 % & $\mathbf{79.34}$ & 74.12 & 76.52\\
%                 % & $\textBF{79.34}$ & 74.12 & 76.52\\
%                 & 79.30 & 73.81 & 76.32\\ % CoCoA-Mix
%             \cmark & \cmark
%                 % & 79.31 & $\mathbf{75.00}$ & $\mathbf{76.94}$\\
%                 % & 79.31 & $\textBF{75.00}$ & $\textBF{76.94}$\\
%                 & $\textBF{79.31}$ & $\textBF{75.10}$ & $\textBF{77.03}$\\ % CoCoA-Mix
%         \bottomrule
%     \end{tabular}
% \end{center}
% \end{table}

\begin{figure}[t!]
\centering
% --- Table (left side) ---
\begin{minipage}[t]{0.40\textwidth}
\centering
\captionof{table}{
Effect of \vtwo{CoA-weight} on {\Base} and {\New} classes.
}
\label{tab:ablation_temp}
\begin{tabular}{c c c c c}
    \toprule 
    \vtwo{$\pi_i^\text{in}$} & \vtwo{$\pi_i^\text{out}$}
    & \MethodFont{Base} & \MethodFont{New} & \MethodFont{H} \\
    \midrule
    \xmark & \xmark & 79.12 & 73.66 & 76.15\\
    \cmark & \xmark & 79.30 & 73.81 & 76.32\\
    \cmark & \cmark & $\textBF{79.31}$ & $\textBF{75.10}$ & $\textBF{77.03}$\\
    \bottomrule
\end{tabular}
\end{minipage}
\hfill
% --- Figure (right side) ---
\begin{minipage}[t]{0.56\textwidth}
\centering
\subfigure[] {
    \includegraphics[width=0.46\linewidth]{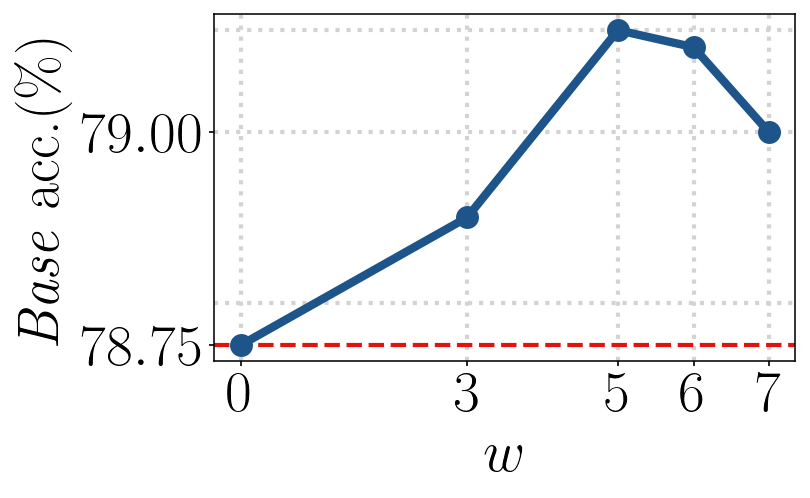}
    \label{fig:sensitivity_a}
}
% \vspace{1ex}
\subfigure[] {
    \includegraphics[width=0.46\linewidth]{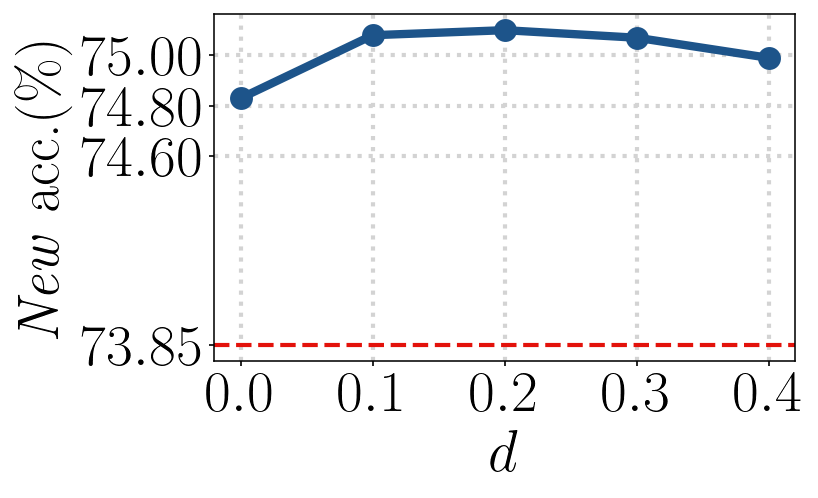}
    \label{fig:sensitivity_b}
}
\caption{
Sensitivity analysis of hyperparameters. The dotted line indicates baseline performance. (a) Effect of CoA-loss weight $w$ on {\Base} accuracy. (b) Effect of margin $d$ on {\New} accuracy.
}
\label{fig:sensitivities}
\end{minipage}

\end{figure}

% \begin{figure}[t]
%     \centering
%     \subfigure[]%[Effect of $w_1$ on \textit{Base} performance]
%     {\label{fig:sensitivity_a}\includegraphics[width=50mm]{resources/sensitivity_base.png}}
%     \subfigure[]%[Effect of $d$ on \textit{New} performance]
%     {\label{fig:sensitivity_b}\includegraphics[width=50mm]{resources/sensitivity_new.png}}    \caption{Sensitivity analysis of hyperparameters. The dotted line represents the baseline performance. (a) Average performance on {\Base} with varying CoA-loss weight $w$. (b) Average performance on {\New} with varying margin $d$.}\label{fig:sensitivities}
% \end{figure}

To evaluate the effectiveness of \vtwo{CoA-weight}, we conducted an ablation study on optimizing \vtwo{weights} for \textit{Base} classes (\vtwo{$\pi_i^\text{in}$}) and \textit{New} classes (\vtwo{$\pi_i^\text{out}$}) separately and jointly.
\cref{tab:ablation_temp} shows that optimizing \vtwo{$\pi_i^\text{in}$} for \textit{Base} classes improves performance over the baseline.
This indicates that confidence adjustment in the in-class domain provides additional benefits for specialization.
When both \vtwo{$\pi_i^\text{in}$ and $\pi_i^\text{out}$} are optimized, the performance improves further, achieving the best harmonic mean between the \textit{Base} and \textit{New} classes. These results demonstrate that adapting \vtwo{weights} for both in-classes and out-classes is crucial for achieving generalization without compromising specialization.
% \EDITED{More ablation studies can be found in~\cref{sec:appendix-ablation_studies}.}

% === % 
\subsection{Sensitivity of Hyperparameters}

We analyzed the sensitivity of our method to hyperparameters by evaluating performance under different configurations. 
Specifically, we investigated the effect of varying weights $w$ for CoA-loss on \textit{Base} classes and the margin $d$ for CoA-weights on \textit{New} classes. 

\cref{fig:sensitivity_a} shows the \textit{Base} performance for different $w$ values, with the red dashed line representing the standard cross-entropy baseline.
Introducing CoA-loss improves specialization, and tuning $w$ effectively balances the focus on confusing cases, achieving superior performance.
\cref{fig:sensitivity_b} shows the {\New} performance with varying margins $d$. 
The red dashed line represents the baseline without $\pi^\text{out}$.
Across all hyperparameter settings, our method outperforms the baseline, indicating that performance is relatively insensitive to the choice of margin $d$.

% === % 
\subsection{Generation of Unseen Classes}

\begin{table}[b]
\caption{Ablation study comparing different strategies for generating unseen classes. The table reports accuracy on \textit{New} classes.}\label{tab:ablation_generation_unseen}
\begin{center}
    \footnotesize{
    \begin{tabular}{@{}c | c c c c @{}}
        \toprule 
            & \MethodFont{None} & \MethodFont{Random String} & \MethodFont{Random String and Word} & \MethodFont{Random Word}\\
        \midrule
            \MethodFont{Accuracy}
            & 74.12 & 75.00 & 75.04 & $\mathbf{75.10}$\\ 
        \bottomrule
    \end{tabular}
    }
\end{center}
\end{table}

We conducted an ablation study to evaluate the effectiveness of different strategies for generating unseen classes. The study compares three methods: (1) no generation (\texttt{None}), (2) generating random strings (\texttt{Random String}), (3) generating a total of $N$ classes, with half as random strings and half sampled from the random word API~\cite{wonderwords} (\texttt{Random String and Word}), and (4) sampling $N$ words from the random word API~\cite{wonderwords} (\texttt{Random Word}). 
% (3) generating a total of $N$ labels, with half as random strings and half sampled from the random word API~\cite{wonderwords} (\texttt{Random String and Word}),
As represented in~\cref{tab:ablation_generation_unseen}, sampling random words yields the best performance, outperforming other methods.
% This suggests that generating out-classes in a form easily interpretable by CLIP further enhances generalization compared with other strategies.
\EDITED{
This suggests that generating out-classes using semantically meaningful words, rather than arbitrary strings, further enhances generalization.
This occurs because semantically meaningful words better align with learned embeddings, providing clearer semantic cues for distinguishing unseen classes, whereas arbitrary strings introduce noise into the embedding space.
}

\section{Qualitative Visualizations}\label{sec:appendix-qualitative_visualization}

%  using the PyTorch library from Gildenblat and contributors~\cite{jacobgilpytorchcam},
% We provide ScoreCAM~\cite{wang2020score} visualizations using~\cite{jacobgilpytorchcam} comparing CoCoA-Mix against zero-shot CLIP and CoOp to better understand model behavior. 
To better understand model behavior, we provide ScoreCAM~\cite{wang2020score} using the PyTorch library by Gildenblat et al.~\yrcite{jacobgilpytorchcam}, comparing CoCoA-Mix with zero-shot CLIP and CoOp. 
The visualization results are shown in~\cref{fig:appendix-scorecam}. In the \textit{Flowers102} dataset (\textit{New}), CoCoA-Mix more accurately attends to semantically meaningful regions in out-class samples, suggesting that CoA-weights effectively enhance generalization. In the \textit{FGVC Aircraft} dataset (\textit{Base}), CoA-loss focuses more precisely on fine-grained details such as text on airplane wings, outperforming zero-shot CLIP in specialization. These qualitative results demonstrate that CoA-loss and CoA-weights contribute to generalization and specialization, respectively.

\newcommand{\cmarkk}{{\textcolor{Green}{\ding{51}}}}%
\newcommand{\xmarkk}{\textcolor{Red}{\ding{55}}}%
\newcommand\correct[1][Green!100]{%
  \tikz[baseline={(char.base)}]\node[circle,draw=#1, line width=0.8pt, inner sep=0pt](char){\cmarkk};\ %
}
\newcommand\incorrect[1][Red!100]{%
  \tikz[baseline={(char.base)}]\node[circle,draw=#1, line width=0.8pt, inner sep=0pt](char){\xmarkk};\ %
}
\newcommand\Gt[1]{\makecell[cc]{\small #1}}
\newcommand\Gtt[2]{\makecell[cc]{\small #1 \\ \small #2}}
\newcommand\Neg[1]{\makecell[cc]{\incorrect\\\small #1}}
\newcommand\Negg[2]{\makecell[cc]{\incorrect\\\small #1 \\\small #2}}
\newcommand\Pos[1]{\makecell[cc]{\correct\\\small #1}}
\newcommand\Poss[2]{\makecell[cc]{\correct\\\small #1 \\ \small #2}}
\newcommand\methodOne[1]{\makecell[cc]{\small\textbf{#1}}}
\newcommand\methodTwo[2]{\makecell[cc]{\small\textbf{#1}\\\small\textbf{#2}}}

\setlength{\fboxsep}{0pt}   % 이미지와 테두리 사이 여백 제거
\setlength{\fboxrule}{2pt}  % 테두리 두께 조절
\begin{figure}[H]\centering
    \begin{tabular}{rcccccc}
        
        & \methodOne{Image} 
        & \methodOne{Zero-Shot CLIP} 
        & \methodOne{CoOp}
        & \methodTwo{CoA-Loss}{(w/ Ensemble)}
        & \methodOne{CoCoA-Mix}
        \\[6pt]
        \raisebox{2.2em}{\rotatebox[origin=c]{90}{\textbf{Flowers102 (Base)}}}
        & \includegraphics[width=0.12\linewidth,height=\textheight,keepaspectratio]{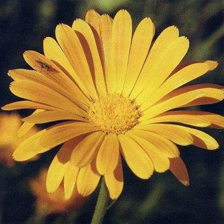} 
        & \adjustbox{cfbox=Red 2pt 0pt}{{\includegraphics[width=0.12\linewidth,height=\textheight,keepaspectratio]{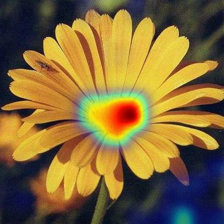}}}
        & 
        \adjustbox{cfbox=Red 2pt 0pt}{\includegraphics[width=0.12\linewidth,height=\textheight,keepaspectratio]{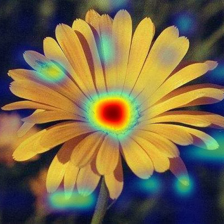}}
        & 
        \adjustbox{cfbox=Green 2pt 0pt}{\includegraphics[width=0.12\linewidth,height=\textheight,keepaspectratio]{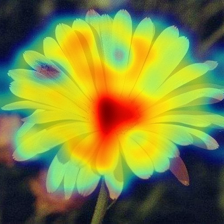}}
        & 
        \adjustbox{cfbox=Green 2pt 0pt}{\includegraphics[width=0.12\linewidth,height=\textheight,keepaspectratio]{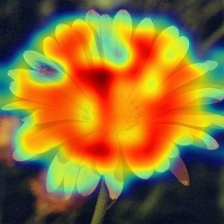}}
        \\[-10pt]
        & \Gt{english marigold}
        & \Gt{mexican aster}
        & \Gt{barbeton daisy}
        & \Gt{english marigold}
        & \Gt{english marigold}
    % ---------------------------------------- % 
        \\[15pt]
        \raisebox{2.2em}{\rotatebox[origin=c]{90}{\textbf{Flowers102 (New)}}}
        & \includegraphics[width=0.12\linewidth,height=\textheight,keepaspectratio]{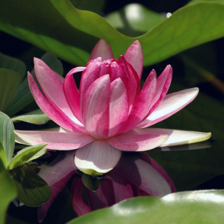} 
        & 
        \adjustbox{cfbox=Red 2pt 0pt}{\includegraphics[width=0.12\linewidth,height=\textheight,keepaspectratio]{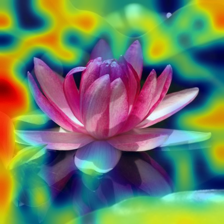}}
        & 
        \adjustbox{cfbox=Red 2pt 0pt}{\includegraphics[width=0.12\linewidth,height=\textheight,keepaspectratio]{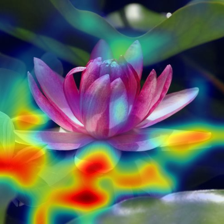}}
        &         
        \adjustbox{cfbox=Red 2pt 0pt}{\includegraphics[width=0.12\linewidth,height=\textheight,keepaspectratio]{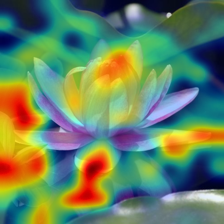}}
        & 
        \adjustbox{cfbox=Green 2pt 0pt}{\includegraphics[width=0.12\linewidth,height=\textheight,keepaspectratio]{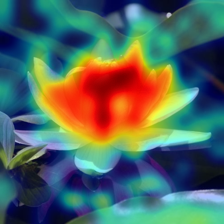}}
        \\[-10pt]
        & \Gt{water lily}
        & \Gt{lotus}
        & \Gt{lotus}
        & \Gt{lotus}
        & \Gt{water lily}
    
    % ---------------------------------------- % 
        \\[15pt]
        \raisebox{2.2em}{\rotatebox[origin=c]{90}{\textbf{Flowers102 (New)}}}
        & \includegraphics[width=0.12\linewidth,height=\textheight,keepaspectratio]{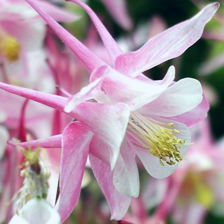} 
        & 
        \adjustbox{cfbox=Red 2pt 0pt}{\includegraphics[width=0.12\linewidth,height=\textheight,keepaspectratio]{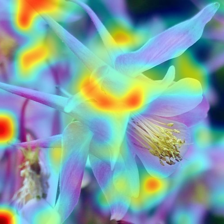}}
        & 
        \adjustbox{cfbox=Red 2pt 0pt}{\includegraphics[width=0.12\linewidth,height=\textheight,keepaspectratio]{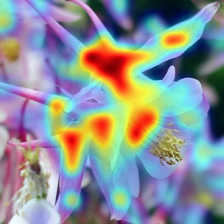}}
        & 
        \adjustbox{cfbox=Green 2pt 0pt}{\includegraphics[width=0.12\linewidth,height=\textheight,keepaspectratio]{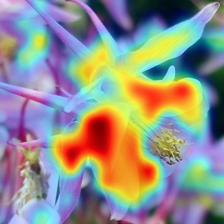}}
        & 
        \adjustbox{cfbox=Green 2pt 0pt}{\includegraphics[width=0.12\linewidth,height=\textheight,keepaspectratio]{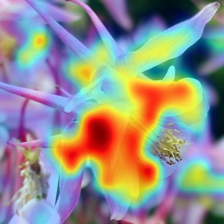}}
        \\[-10pt]
        & \Gt{columbine}
        & \Gt{gaura}
        & \Gt{bee balm}
        & \Gt{columbine}
        & \Gt{columbine}
    
    % ---------------------------------------- % 
        \\[15pt]
        \raisebox{2.2em}{\rotatebox[origin=c]{90}{\textbf{FGVCAircraft (Base)}}}
        & \includegraphics[width=0.12\linewidth,height=\textheight,keepaspectratio]{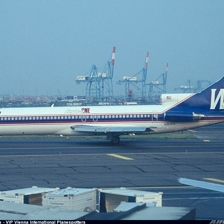} 
        & 
        \adjustbox{cfbox=Red 2pt 0pt}{\includegraphics[width=0.12\linewidth,height=\textheight,keepaspectratio]{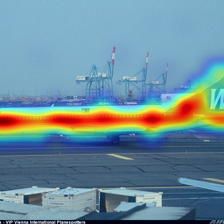}}
        & 
        \adjustbox{cfbox=Red 2pt 0pt}{\includegraphics[width=0.12\linewidth,height=\textheight,keepaspectratio]{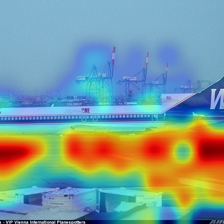}}
        & 
        \adjustbox{cfbox=Green 2pt 0pt}{\includegraphics[width=0.12\linewidth,height=\textheight,keepaspectratio]{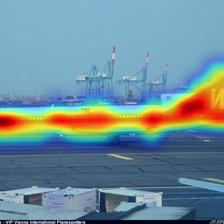}}
        & 
        \adjustbox{cfbox=Green 2pt 0pt}{\includegraphics[width=0.12\linewidth,height=\textheight,keepaspectratio]{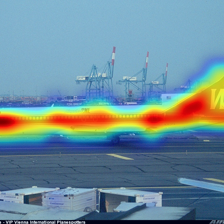}}
        \\[-10pt]
        & \Gt{727-200}
        & \Gt{Boeing 717}
        & \Gt{707-320}
        & \Gt{727-200}
        & \Gt{727-200}
    
    % ---------------------------------------- % 
        \\[15pt]
        \raisebox{2.2em}{\rotatebox[origin=c]{90}{\textbf{FGVCAircraft (New)}}}
        & \includegraphics[width=0.12\linewidth,height=\textheight,keepaspectratio]{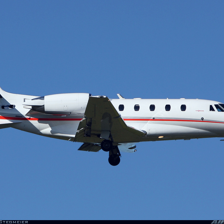} 
        & 
        \adjustbox{cfbox=Red 2pt 0pt}{\includegraphics[width=0.12\linewidth,height=\textheight,keepaspectratio]{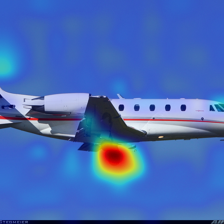}}
        & 
        \adjustbox{cfbox=Red 2pt 0pt}{\includegraphics[width=0.12\linewidth,height=\textheight,keepaspectratio]{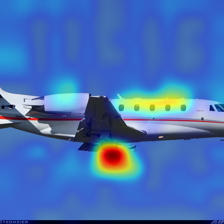}}
        & 
        \adjustbox{cfbox=Green 2pt 0pt}{\includegraphics[width=0.12\linewidth,height=\textheight,keepaspectratio]{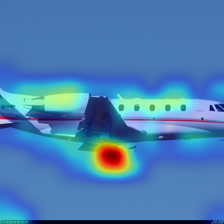}}
        & 
        \adjustbox{cfbox=Green 2pt 0pt}{\includegraphics[width=0.12\linewidth,height=\textheight,keepaspectratio]{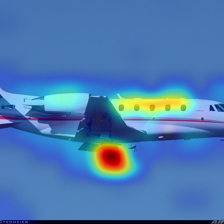}}
        \\[-10pt]
        & \Gt{Cessna 560}
        & \Gt{Embraer Legacy 600}
        & \Gt{Falcon 2000}
        & \Gt{Cessna 560}
        & \Gt{Cessna 560}
    \end{tabular}

    \caption{ScoreCAM results on Flowers102 and FGVCAircraft. Red indicates higher activation. The leftmost column shows the image and ground-truth label. Other columns show the ScoreCAM results and predicted labels for each method.}\label{fig:appendix-scorecam}
\end{figure}

\newpage
\section{Limitation and Future Work}\label{sec:appendix-limitations}

This study focused on textual prompt tuning.
To demonstrate that improved performances in the source domain lead to better performance in the target domain, we adopted the following relation from Nguyen et al.~\yrcite{nguyen2022kl}:
\begin{equation}
    \err{T}{\pred} 
    \leq \err{S}{\pred} +\frac{C}{\sqrt{2}}\sqrt{
        \KLD{p_T(\bz)}{p_S(\bz)} + \delta
    },
\end{equation}
\noindent where $\bz$ is defined as a visual embedding; $C$ is a constant that bounds $\log{\pred(l)}$, ensuring each class probability is at least $\exp(-C)$; 
$p_T$ and $p_S$ are the marginal distribution of $\bz$ for the target and source domains, respectively;

This assumes that visual embeddings $\bz$ remain consistent across domains, implying similar embeddings for images of the same semantic but differing styles.
However, pre-trained CLIP can generate distinct visual embeddings for images with varying styles.
As a result, while the proposed method ensures that prompts specialized for the source domain perform well in the target domain from a class perspective, it does not guarantee performance across target domains with different styles.
This limitation is reflected in the results, where cross-dataset transfer shows weaker generalization compared with base-to-new generalization.
Future work could address this by optimizing visual prompts to align embeddings of images with different styles but the same semantics.
Such methods could enhance specialization and generalization across both classes and styles, enabling more robust performance across diverse domains.

\end{document}